\documentclass[letterpaper, 10 pt, journal, twoside]{IEEEtran}  % Comment this line out
\IEEEoverridecommandlockouts                              % This command is only
\usepackage{graphicx}
\usepackage{siunitx}
\usepackage[utf8]{inputenc}
\usepackage[T1]{fontenc}
\usepackage{amsfonts}
\newcommand{\fig}[1]{Fig.~\ref{#1}}
\newcommand{\tab}[1]{Table~\ref{#1}}
\newcommand{\eq}[1]{(\ref{#1})}
\newcommand{\bm}[1]{{\mbox{\boldmath $#1$}}}
\usepackage{epsfig} % for postscript graphics files
\usepackage{amsmath} % assumes amsmath package installed
\usepackage{subcaption}
\usepackage{caption}
\usepackage{color}
\title{Risk-Aware Motion Planning for a Limbed  Robot  with  Stochastic Gripping Forces Using Nonlinear Programming}

\author{Yuki Shirai$^{1}$, Xuan Lin$^{1}$, Yusuke Tanaka$^{1}$, Ankur Mehta$^{2}$, and Dennis Hong$^{1}$% <-this % stops a space
\thanks{Manuscript received: February, 24, 2020; Revised May, 8, 2020; Accepted May, 30, 2020.}%Use only for final RAL version
\thanks{This paper was recommended for publication by Editor Abderrahmane Kheddar upon evaluation of the Associate Editor and Reviewers' comments. 
(\emph{Corresponding author: Yuki Shirai.})
This work was partially supported by ONR through a grant N00014-15-1-2064. The work of Y. Shirai was partially supported by the Funai Foundation for Information Technology. 
}%Use only for final RAL version
\thanks{$^{1}$Y. Shirai, X. Lin, Y. Tanaka, and D. Hong are with the Robotics and Mechanisms Laboratory, Department of Mechanical and Aerospace Engineering, University of California, Los Angeles, CA 90095, USA.
        {\tt\small yukishirai4869@g.ucla.edu}, {\tt\small maynight@ucla.edu}, {\tt\small yusuketanaka@g.ucla.edu}, {\tt\small dennishong@ucla.edu}
        }
\thanks{$^{2}$A. Mehta is with the Laboratory for Embedded Machines and Ubiquitous Robotics, Department of Electrical and Computer Engineering, University of California, Los Angeles, CA 90095, USA.
        {\tt\small mehtank@ucla.edu}
        }%        
}

\IEEEpubid{\begin{minipage}{\textwidth}\ \\[12pt] \centering
\textcopyright  2020 IEEE. Personal use of this material is permitted.  Permission from IEEE must be obtained for all other uses, in any current or future media, including reprinting/republishing this material for advertising or promotional purposes, creating new collective works, for resale or redistribution to servers or lists, or reuse of any copyrighted component of this work in other works.
\end{minipage}} 

\begin{document}
% \IEEEpubid{978-1-5386-2880-5/17/\$31.00~\copyright~2017 IEEE}
\maketitle
% \thispagestyle{empty}
% \pagestyle{empty}
%%%%%%%%%%%%%%%%%%%%%%%%%%%%%%%%%%%%%%%%%%%%%%%%%%%%%%%%%%%%%%%%%%%%%%%%%%%%%%%%
%%note: must be manage risk in a "quantitative" way! Model and environmental uncertainty are coupled.
%% reduce to simpler deterministic optimization when I assume the noise follows gauusian

% Paper headers 
\markboth{IEEE Robotics and Automation Letters. Preprint Version. ACCEPTED  JUNE, 2020}
{SHIRAI \MakeLowercase{\textit{et al.}}: Risk-Aware Motion Planning for a Limbed  Robot  with Stochastic Gripping Forces Using Nonlinear Programming} 

\begin{abstract}
We present a motion planning algorithm with probabilistic guarantees for limbed robots with \textcolor{black}{stochastic gripping forces}.  \textcolor{black}{Planners based on deterministic models with a worst-case uncertainty can be conservative and inflexible} to \textcolor{black}{consider} the stochastic behavior of the contact, especially when a gripper is installed.  \textcolor{black}{Our} proposed planner enables the robot to simultaneously plan its pose and contact force trajectories \textcolor{black}{while considering} the risk associated with the gripping forces. \textcolor{black}{Our} planner is formulated as a nonlinear programming problem with chance constraints, which allows the robot to generate a variety of motions based on different risk bounds.
To model the gripping forces as random variables, we employ Gaussian Process regression. We validate \textcolor{black}{our proposed} motion planning algorithm on an 11.5 kg six-limbed robot for two-wall climbing. Our results show that \textcolor{black}{our} proposed planner generates various trajectories (e.g., avoiding low friction terrain under the  low risk bound, choosing an unstable but faster gait under the high risk bound) by changing the probability of risk based on \textcolor{black}{various specifications}.
\end{abstract}
\begin{IEEEkeywords} Legged Robots, Motion and Path Planning,  Optimization and Optimal Control
 \end{IEEEkeywords}

%%%%%%%%%%%%%%%%%%%%%%%%%%%%%%%%%%%%%%%%%%%%%%%%%%%%%%%%%%%%%%%%%%%%%%%%%%%%%%%%
\section{INTRODUCTION}
\IEEEPARstart{P}{lanning} complex motions for limbed robots is challenging because planners need to design footsteps and a body trajectories while considering the robot  kinematics and  reaction forces.
Motion planning for limbed robots has been studied by a number of researchers. Sampling-based planning, such as the Probabilistic-Roadmap (PRM), samples the environment and generates the motion while satisfying static equilibrium and kinematics for a robot \cite{PRM}, \cite{RRT}. Optimization-based planning, such as Mixed-Integer Convex Programming (MICP) and Nonlinear Programming (NLP),  solves the solution given constraints using optimization algorithms such as gradient descent \cite{MIT_envelope}-\cite{xuanMICP}.

While many papers discuss motion planning for the robot, few studies have investigated how planning is affected by \textcolor{black}{stochastic gripping forces}.
One of the open problems in motion planning of a limbed robot equipped with grippers is the stochastic nature of gripping \cite{linear_gripper}, \cite{linear_gripper2}. \textcolor{black}{For example,} the gripping forces caused by spine grippers depend on the stochastically distributed asperity strength \textcolor{black}{(\fig{robot_wall})}. Thus, \textit{risk} results from the randomness of the gripping force. By considering risk in a probabilistic manner, the planner \textcolor{black}{can} design a variety of trajectories based on \textcolor{black}{various specifications}. 

   \begin{figure}[!t]
    \centering
\includegraphics[width=0.42\textwidth, clip]{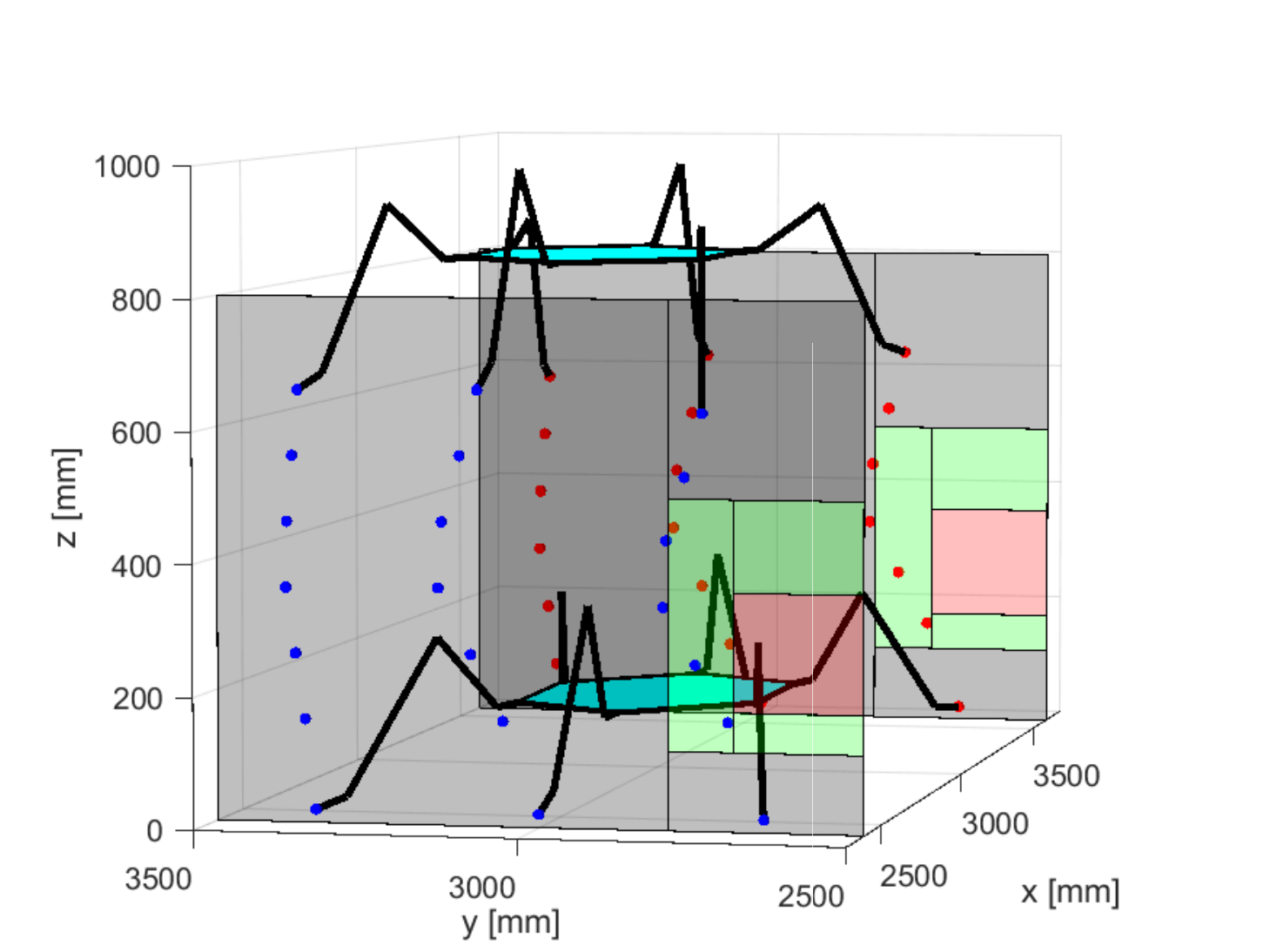}
    \caption{\small A planned trajectory for wall climbing \textcolor{black}{that considers risk arising from slippery terrain.} The black area shows a high friction area, the green area shows a low friction area, and the red area shows a zero friction area. Blue and red dots show the planned foot positions, and the hexagons show the body of the robot.}
    \label{low_friction_avoid_a_3d}
\end{figure}

% boundary vs density function
The stochastic planning problem can be categorized into two approaches: robust approaches \textcolor{black}{\cite{robust_gait}-\cite{robust_torque}, \cite{autocar}} and risk-bounded approaches \textcolor{black}{\cite{robust_torque}, \cite{takanishi}, \cite{blackmore2}-\cite{rover}}. In robust approaches, the planners design trajectories that guarantee the feasibility of the motion given the uncertainty bounds.
\textcolor{black}{A soft constraints-based robust planning was investigated in \cite{winkler_zmp}, where the planner allows the solution to be at the boundary of stability. Tas showed the planner to remain collision-free for the worst-case uncertainty for automated driving \cite{autocar}.}

\IEEEpubidadjcol

 On the other hand, the risk-bounded approach designs trajectories that guarantee the feasibility of the motion given the probability density function (PDF): it prevents the probability of violating state constraints (violation probability) from being higher than a pre-specified probability.
 \textcolor{black}{Prete formulated a chance-constrained optimization problem of a bipedal robot by approximating a joint chance constraints with linear inequality constraints \cite{robust_torque}. Planning on slippery terrain was in \cite{takanishi}, where the planner utilizes the prediction of the coefficient of friction to design the motion of the body and footsteps, respectively. Our approach is similar to \cite{takanishi}, but we model the stochastic contact force of the robot and formulate the planning algorithm considering the trajectory of a body and footsteps  simultaneously.} 

\textcolor{black}{For tasks with a higher probability of failure (e.g., climbing on slippery terrain) \cite{rover},}
the risk-bounded approach has advantages over the robust approach. Because the robust approach
\textcolor{black}{often uses a much less informative deterministic model, it is likely to generate conservative solutions with the worst-case uncertainty bound. For demanding tasks, this may be infeasible, with such a planner generating no possible solution and failing to achieve specified goals. In contrast, because the risk-bounded approach can be more aggressive, the problem may be feasible, generating trajectories that carry a probability of failure through risk-taking alongside a non-zero chance of successfully achieving the goal.} The violation probability provides a tuning knob
\textcolor{black}{
 to define a Pareto boundary on the risk between failure while finding a trajectory vs. failure while executing a found trajectory. This user-defined parameter can be task- and environment-specific, in contrast to the rigidity of the robust approach.}

In this paper, we \textcolor{black}{address} a motion planning algorithm formulated as NLP for a limbed robot with stochastic gripping forces. \textcolor{black}{Our proposed} planner solves for stable postures and forces simultaneously with guaranteed bounded risk. In addition, chance constraints are introduced into the planner that restrict contact forces in a probabilistic manner. We employ a Gaussian Process (GP), a non-parametric Bayesian regression tool, to acquire the PDF of the gripping force. \textcolor{black}{Our proposed} motion planning algorithm is validated on an 11.5 kg hexapod robot with spine grippers for multi-surface climbing.
\textcolor{black}{While we focus on multi-surface robotic climbing with spine grippers in this paper},
\textcolor{black}{our proposed} planner can be applied to other robots with any \textcolor{black}{type of} grippers for \textcolor{black}{performing any task}  (e.g., planning of walking, grasping) as long as the robot has contact points with stochastic models.

The contributions of this \textcolor{black}{paper are} as follows:
\begin{enumerate}
    \item We formulate risk-bounded NLP-based planning that considers the stochasticity of gripping forces.
    % \item We include gripping forces by spine grippers as random variables in the planner and formulates the chance constrained motion planner 
    \item We employ the Gaussian Process to model gripping forces as random variables.
    \item We validate the algorithm in hardware experiments.
\end{enumerate}

\section{PROBLEM FORMULATION}\label{sec:pf}
This section describes the \textcolor{black}{friction cone considering maximum gripping forces,} model of a position-controlled limbed robot with multi-contact surfaces, \textcolor{black}{and} the modeling process of a gripping force through GP.

\subsection{Friction Cone with Stochastic Gripping Forces}\label{friction_cone}
\begin{figure}
    \centering
    \includegraphics[width=0.34\textwidth, clip]{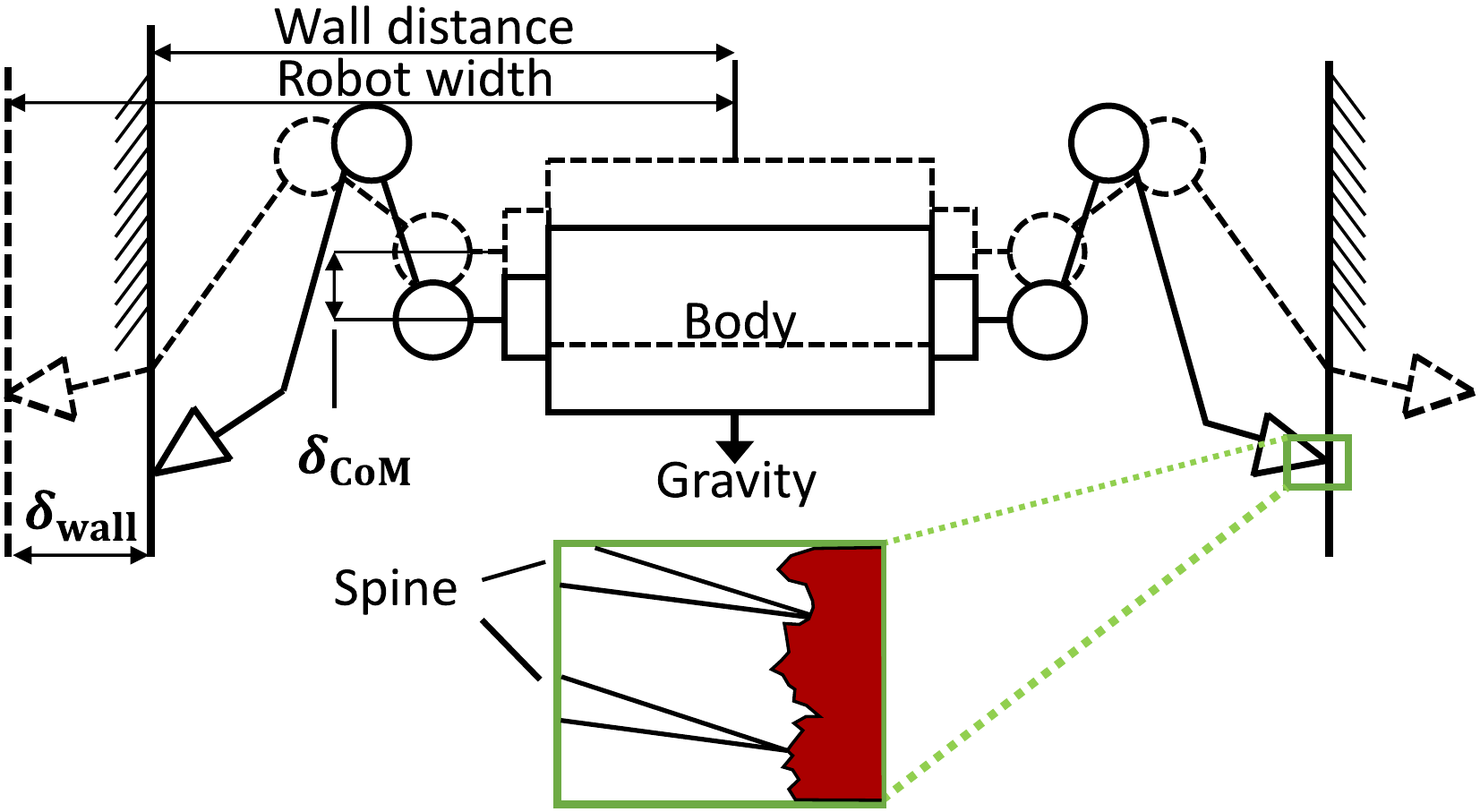}
    \caption{Deflection of a multi-limbed robot bracing between walls}
    \label{robot_wall}
\end{figure}

\textcolor{black}{With grippers, the friction cone constraint can be relaxed on the contact point.}
% In our work, the robot is equipped with spine grippers.
For our spine-based gripper, even under a zero normal load, the spines insert into the microscopic gaps on the surface (\fig{robot_wall}), generating a significant amount of shear force (\fig{GP_result}) \textcolor{black}{\cite{nagaoka}}. 
\textcolor{black}{For a magnet-based gripper, the reaction forces includes the additional magnetic force imposed by the gripper itself, offsetting the friction cone as seen by the rest of the robot.}

\begin{figure}
    \centering
    \includegraphics[width=0.27\textwidth, clip]{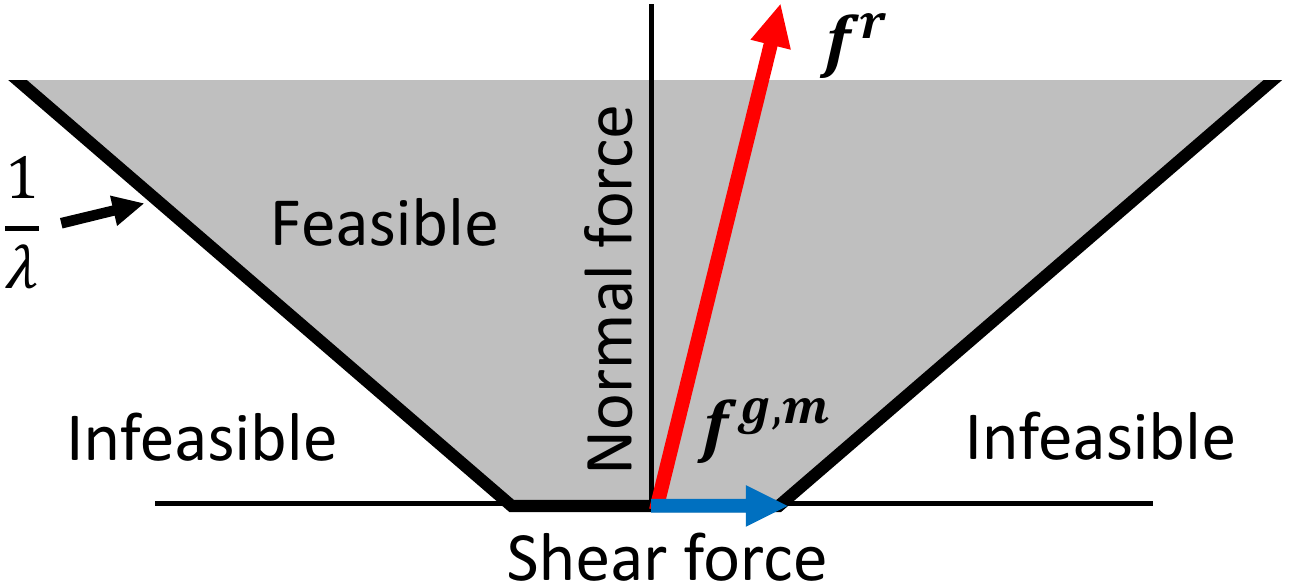}
    \caption{\textcolor{black}{Friction cone considering stochastic gripping forces}}
    \label{friction_setting}
\end{figure}

\textcolor{black}{
Thus, we modify the regular friction cone, adding in an offset shear force when a normal force is zero to account for the gripping force. As the normal force increases, the maximal allowable shear force increase in the same way as a regular frictional force, with a coefficient of friction $\lambda$ that is assumed to be a constant only depending on the property of the contact surface. This contact model is illustrated in \fig{friction_setting}, where $\bm{ f}^r$ is the reaction force between the surface and the gripper. $\bm{f}^{g,m}$ is the maximum gripping force from grippers under a zero normal force. Note that $\bm{f}^{g,m}$ is measured per gripper as a unit.
\textcolor{black}{In general, $\bm{f}^{g,m}$ can have both  normal and shear components. However,} for our spine grippers, the normal component of $\bm{f}^{g,m}$ is relatively small, so we assume that the gripper generates only shear adhesion.
The gripper does not slip when $\bm{ f}^r$ is within this friction cone, as indicated by the shaded region in  \fig{friction_setting}. 
Since the interaction between the micro-spines and the surface is highly random, $\bm{f}^{g,m}$ is naturally modeled as a  Gaussian random variable. However, the orientation of the spine and the number of spines in contact with the surface also change as the orientation of the gripper changes, which leads to a shift of the mean and standard deviation of $\bm{f}^{g,m}$. We learn this model from data by GP.} \textcolor{black}{With GP, our proposed planner is able to deal with the stochastic nature of gripping taking into account the gripper orientation.}


% Model of Reaction Force Using Limb Compliance
\subsection{\textcolor{black}{Model of Reaction Force Using Limb Compliance}}
% how to say two wall climbing is interesting??
During multi-surface \textcolor{black}{locomotion}, the robot leverages the compliance from its motors \textcolor{black}{in order} to squeeze itself between multi-surfaces, as depicted in \fig{robot_wall}. 
One difficulty multi-limbed robots have is that reaction forces are statically indeterminate \cite{bigK}. \textcolor{black}{Consequently,} reaction forces cannot be determined by static equilibrium equations when the robot supports its weight more than three contact points. Hence, in order to calculate the reaction force under this condition, the deformation of the robotic system should be \textcolor{black}{considered}.

From the standard elasticity theory, $\textcolor{black}{\bm{ f}^r}$  can be described as the spring force using \textcolor{black}{the} Virtual Joint Method \textcolor{black}{\cite{VJM}}:
\begin{equation}
    \textcolor{black}{\bm{ f}^r}=\bm{K}\left({\bm \delta}_{\operatorname{wall}}- {\bm \delta}_{\operatorname{CoM}}\right) \label{spring_f}
\end{equation}
\begin{equation}
    \bm{ K=\left(J k^{-1} J{^{\top}}\right)^{-1}},\ \bm{ k }= diag(k_i),   i=1,\ldots, \textcolor{black}{H} \label{bigKandJ}
\end{equation}
% \begin{equation}
%     \bm{ k }= diag(k_i),   i=1,\ldots,M
% \end{equation}
% \begin{equation}
%     \delta\bm{ X}=\label{deltaX}
% \end{equation}
where \bm{K} is the stiffness matrix for $\textcolor{black}{H}$ degree-of-freedom limb.
$\bm{ k }$ is a diagonal matrix that has $k_i$ diagonal elements, and $k_i$ is the spring coefficients of the position-controlled servos. $\bm J$ is a $3 \times \textcolor{black}{H}$ Jacobian matrix. 
The deflection is imposed by terrain where ${\bm \delta}_{\operatorname{wall}}$ is the displacement of the terrain and $\bm \delta_{\operatorname{CoM}}$ is the body deflection, sag-down, due to \textcolor{black}{the} compliance as shown in \fig{robot_wall}.

\subsection{Model of Gripping Force Using Gaussian Process}
The objective of using GP is to predict the \textcolor{black}{maximum} gripping force \textcolor{black}{$\bm{f}^{g,m}$} in a probabilistic way.

\begin{figure}[t]
    \centering
    \includegraphics[width=0.39\textwidth, clip]{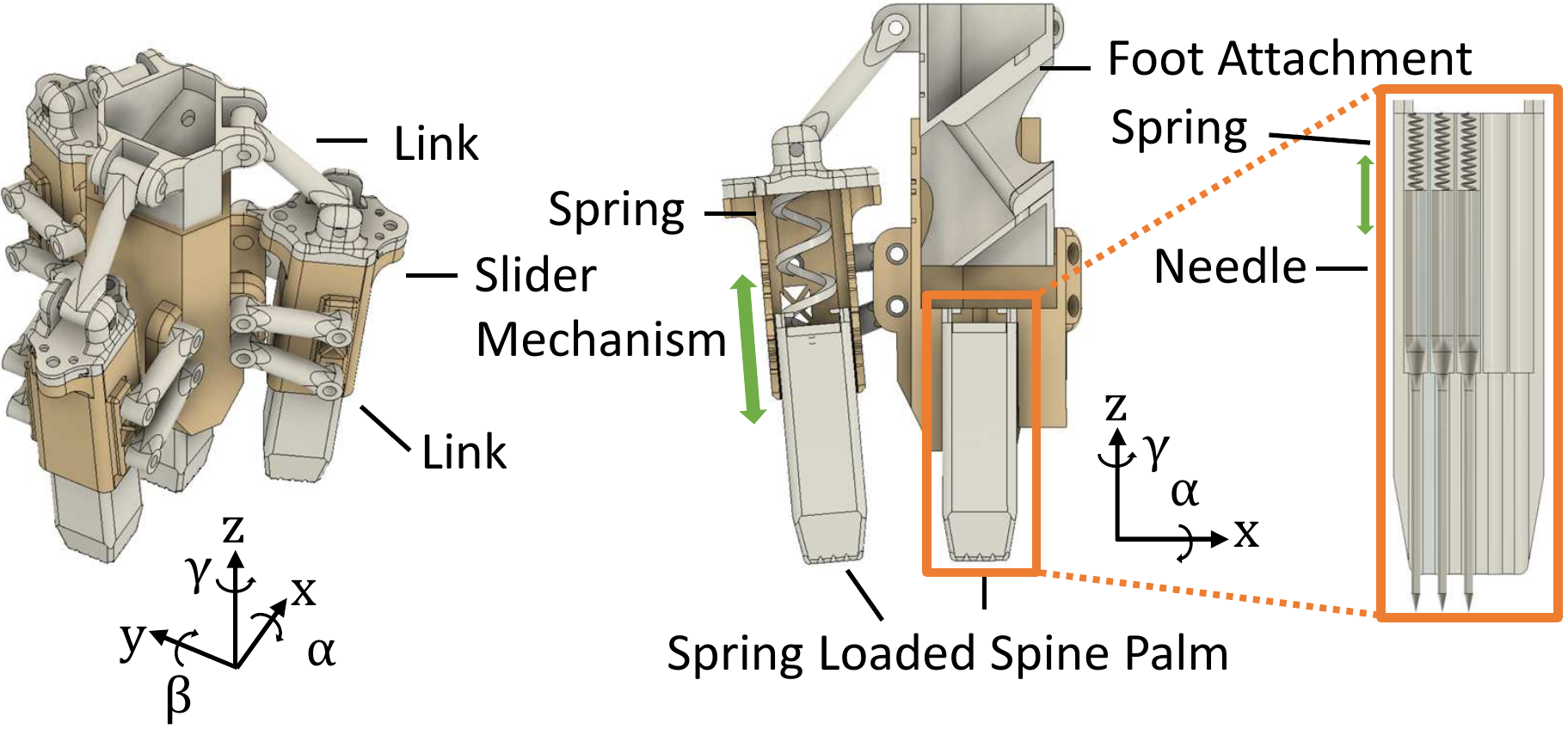}
     \caption{Mechanical design of the spine gripper}\label{CAD}
\end{figure}

\textcolor{black}{There are many design decisions that go into the formulation of the GP problem, including choice of kernel, distance metric, and associated weighting between state variables \cite{GP_text}. We can start with the simplest formulation with all state variables equally weighted under the  Euclidean distance metric using the squared exponential kernel as a starting point. In practice, this choice was observed to work well enough to not necessitate further design. A more general characterization of the effects of these hyperparameters can be found in \cite{GP_text}.}
% We use the SE kernel because the predicted mean converges to zero exponentially outside the boundary region where the training dataset is located. Hence, our NLP-based planner does not explore in the region actively.
In this work, we assume that \textcolor{black}{the maximum} gripping forces by spine grippers is a function of the gripper orientation and the coefficient of friction \textcolor{black}{\cite{linear_gripper}, \cite{linear_gripper2}, \cite{nagaoka}, \cite{microspine}}.  This is because with a microscopic view, \textcolor{black}{the spine-asperity interaction is different depending on how a spine is inclined with respect to the asperity as shown in \fig{robot_wall}.} \textcolor{black}{GP can handle the effects of other unmodeled parameters by treating them into uncertainty.} Hence, the state $\bm{s}$ is a four-dimensional vector with $\bm{s}=[\alpha , \beta , \gamma , \lambda ]^{\top}$ where $\alpha $, $\beta $, $\gamma $ are the \textcolor{black}{Euler} angles along $x$, $y$, $z$ axis defined in \fig{CAD}. 
% $\lambda$ is the coefficient of friction of the surface. 

Here, we assume that the \textcolor{black}{maximum} shear force follows Gaussian distribution. Given a data set $S=\left\{\bm s_{1}, \cdots, \bm s_{n}\right\}$ with the measured shear forces \textcolor{black}{$\bm y^{g,m} = \left[y^{g,m}_1, \ldots, y^{g,m}_n\right]^{\top}$}, the \textcolor{black}{maximum} shear force \textcolor{black}{$\bm f^{g,m}$} by a gripper can therefore be modeled as:
\begin{equation}
 \textcolor{black}{\bm f^{g,m}(\bm{s})}\sim \mathcal{GP}(\bm \mu^{g,m}(\bm{s}), \bm \kappa^{g,m}(\bm{s}, \bm{s_*}))
\end{equation}
where, \textcolor{black}{$\bm{f}^{g,m}=\left[f^{g,m}_1, \ldots, f^{g,m}_n\right]^{\top}$}, $n$ is the number of samples from a GP. 
\textcolor{black}{$\bm \mu^{g,m}(\bm{s}) = \left[\mu^{g,m}_1(\bm s), \ldots, \mu^{g,m}_n(\bm s)\right]^{\top}$} 
is the mean and 
$\bm [\textcolor{black}{\bm \kappa^{g,m}]_{i,j}}=\textcolor{black}{\kappa^{g,m}}\left(\bm{s}_i, \bm{s}_j\right)$ is the covariance matrix, where $\kappa^{g,m}(\cdot, \cdot)$ is a positive definite kernel. In this work, we employ the squared exponential kernel as follows:
\begin{equation}
  \textcolor{black}{\kappa^{g,m}}\left(\bm{s}_{i}, \bm{s}_{j}\right)=\sigma_{f}^{2} \exp \left(-\frac{1}{2} \frac{\left|\bm{s}_{i}-\bm{s}_{j}\right|^{2}}{\ell^{2}}\right)
\end{equation}
where $\sigma_{f}^2$ represents the amplitude parameter and $l$ defines the smoothness of the function \textcolor{black}{$\bm f^{g,m}$}.

Here, let $\bm D=[\bm s_{1}, \cdots, \bm s_{n}]^{\top}$ be the matrix of the inputs. In order to predict the mean and variance matrix at $\bm D_*$, we obtain the predictive mean and variance of the \textcolor{black}{maximum} shear force by assuming that it is jointly Gaussian as follows: 
\begin{equation}
   \textcolor{black}{\hat{\bm f}^{g,m}} =\mathbb{E}\left[\bm f^{g,m}\left(\bm{D}_{*}\right)\right]=\bm{\kappa}_{*}^{\top}\left(\bm{K}_D+\sigma_{n}^{2} \bm{I}\right)^{-1} \bm{y}_g
\end{equation}
%
%
% Variance equation
\begin{equation}\label{variance}
\textcolor{black}{\hat{\bm \Sigma}^{g,m}}=
\mathbb{V}\left[\bm f^{g,m}\left(\bm{D}_{*}\right)\right]=\bm{\kappa}_{* *}-\bm{\kappa}_{*}^{\top}\left(\bm{K}_D+\sigma_{n}^{2} \bm{I}\right)^{-1} \bm{\kappa}_{*}
\end{equation}
where $\bm{\kappa}_{*}=\bm \kappa^{g,m}\left(D_*, D\right)$, $\bm K_D=\bm \kappa^{g,m}\left(D, D\right)$, $\bm{\kappa}_{**}=\bm \kappa^{g,m}\left(D_*, D_*\right)$, and $\sigma_{n}^{2}$ is the variance of the Gaussian observation noise with zero mean. 

\textcolor{black}{Our GP procedure can be generalizable to model other gripping forces as long as the gripping force changes continuously as the orientation of the gripper changes. For instance, the GP approach can be used to model the gecko gripper force \cite{gecko} using the detachment angle as the state of the GP.}

% the detachment angle has an effect on the gripping force by the gecko grippers .

    \begin{figure}[t]
    \centering
    \includegraphics[width=0.3\textwidth, clip]{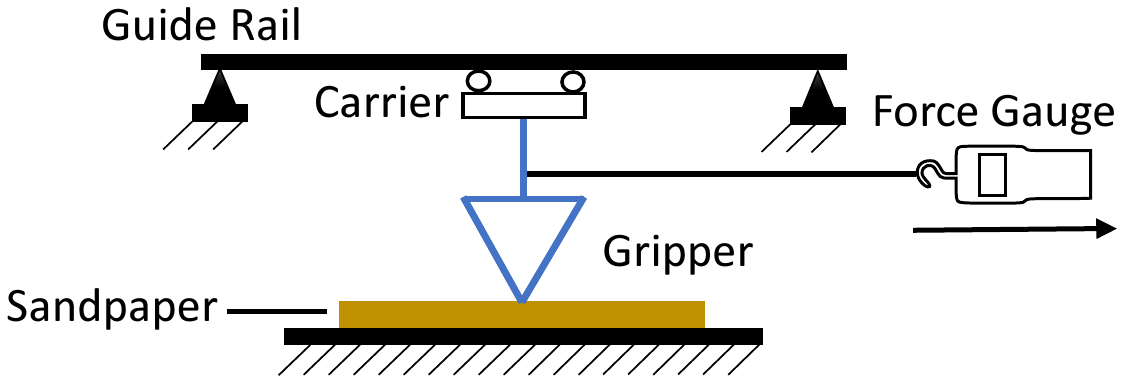}
    \caption{Experiment setup to evaluate \textcolor{black}{maximum} gripping forces on sandpaper}
    \label{experiment_gripper}
\end{figure}

%Theory
% experimental setup

\subsection{Spine Gripper for Wall Climbing}
 A three-finger spine-based gripper \textcolor{black}{was} designed (\fig{CAD}) \textcolor{black}{using} spine cells based on \cite{microspine}.  Each finger consists of a spine cell with 25 machine needles loaded \textcolor{black}{with} 5 mN/mm springs, and a slider mechanism holds the cell with one compliant plastic spring. The diameter of the needle at the tip is 0.93 mm, and it is made of carbon steel. The gripper center module includes one spine palm with the same spine configurations as \textcolor{black}{the} cells.  The attachment component is fixed at the tip of the robot limb at  \ang{37} from the limb axis to maximize the contact area.  \textcolor{black}{The} finger, center, and attachment members are assembled with \textcolor{black}{a one-slider, two linkage mechanism (\fig{CAD}).}
%  two linkage mechanisms with one slider as in \fig{CAD}.
 \textcolor{black}{This} linkage system is designed to provide a passive micro grip as the center palm \textcolor{black}{presses up against a wall}.  The three fingers are located at  \ang{120} apart from each other in $z$-axis and tilted about  \ang{15} from $z$-axis.  
%  The finger slider structure allows the gripper to adapt various wall contact angles and rough surface terrains. 

 \subsection{Data Collection}
% \subsubsection{\textcolor{black}{Training}}
% put figure
\begin{table}[t]
    \caption{\textcolor{black}{Varied orientations for collecting datasets of GP}}
    \centering
    \setlength\tabcolsep{1.5pt}
    \begin{tabular}{c|c|}
      Training& $\alpha,\beta=  \ang{-15},  \ang{0},    \ang{15}$, $\gamma=\ang{0},  \ang{30},  \ang{60}$, $\lambda=1.1, 2.3$ \\
         \hline
         Testing & $\alpha= \ang{-15}, \ang{-10}, \cdots, \ang{15}|\{\beta=\ang{-15}, \gamma=\ang{30}, \lambda=2.3$\}\\
          & $\beta= \ang{-15}, \ang{-10}, \cdots, \ang{15}|\{\alpha=\ang{-15}, \gamma=\ang{30}, \lambda=2.3$\}\\
           & $\gamma= \ang{0}, \ang{15}, \ang{30}, \cdots, \ang{60}|\{\alpha=\ang{-15}, \beta=\ang{-15}, \lambda=2.3$\}\\
           & $\lambda= 1.1,1.4,1.82,2.3|\{\alpha=\ang{0}, \beta=\ang{0}, \gamma=\ang{0}$\}
    \end{tabular}
    \label{GP_dataset}
\end{table}
   \begin{figure}[t]
    \begin{subfigure}{0.24\textwidth}
    \centering
\includegraphics[width=0.9\textwidth, clip]{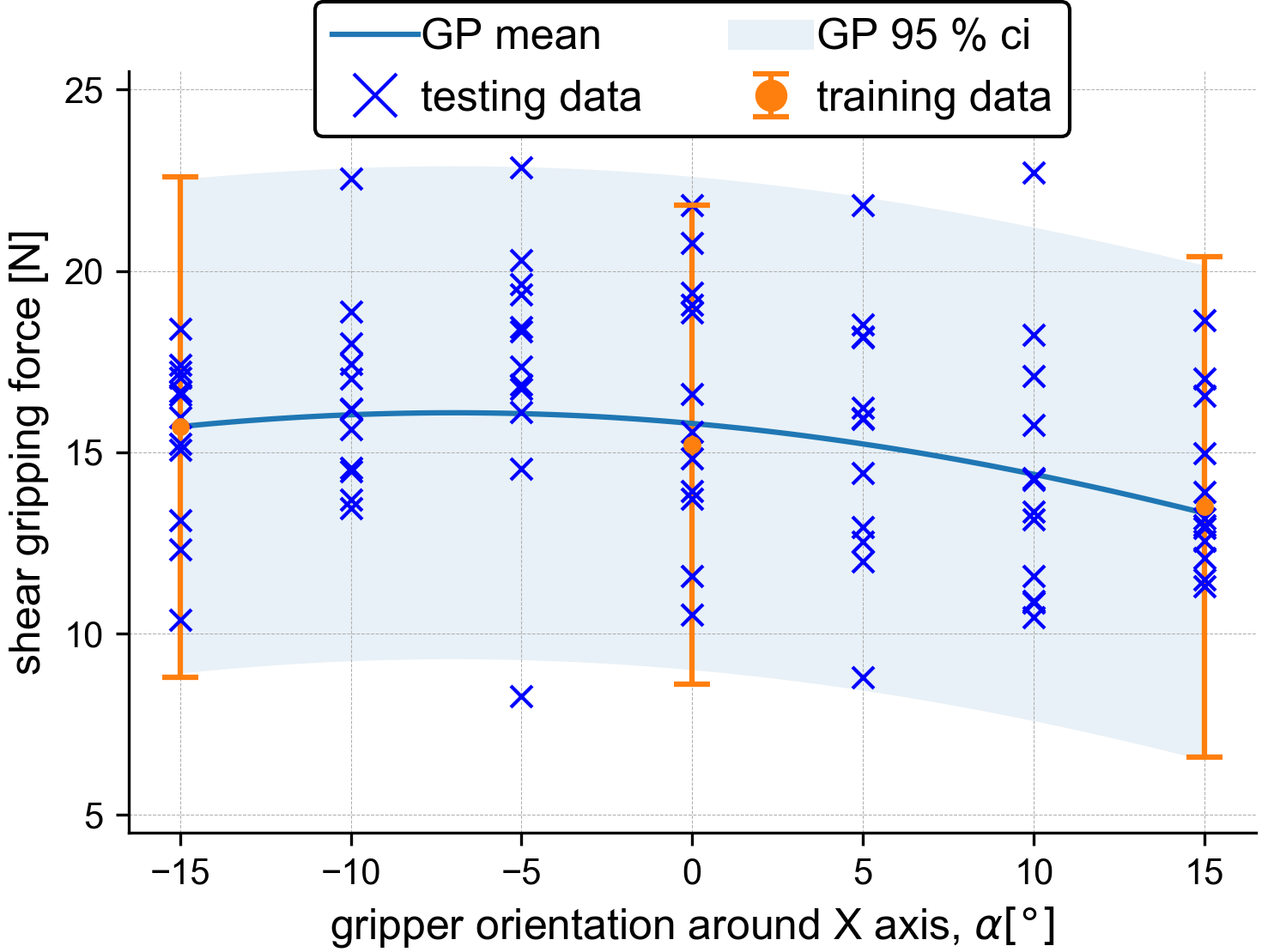}
    \caption{$\beta = \ang{-15}, \gamma = \ang{30}, \lambda = 2.3$}
    \label{GP_x}
    \end{subfigure}
     \begin{subfigure}{0.24\textwidth}
         \centering
    \includegraphics[width=0.9\textwidth, clip]{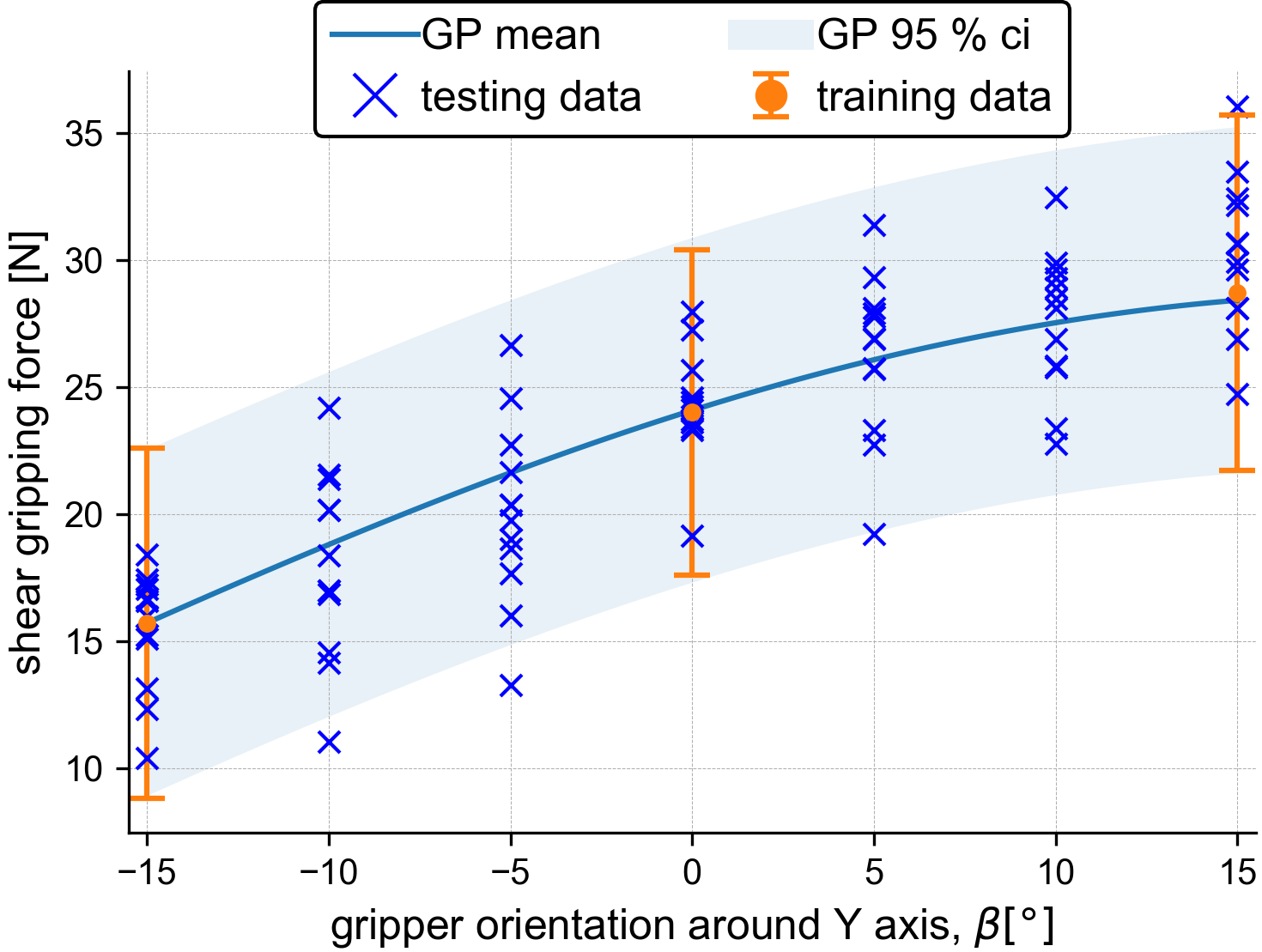}
    \caption{$\alpha = \ang{-15}, \gamma = \ang{30}, \lambda = 2.3$}
    \label{GP_y}
     \end{subfigure}\\
          \begin{subfigure}{0.24\textwidth}
         \centering
    \includegraphics[width=0.9\textwidth, clip]{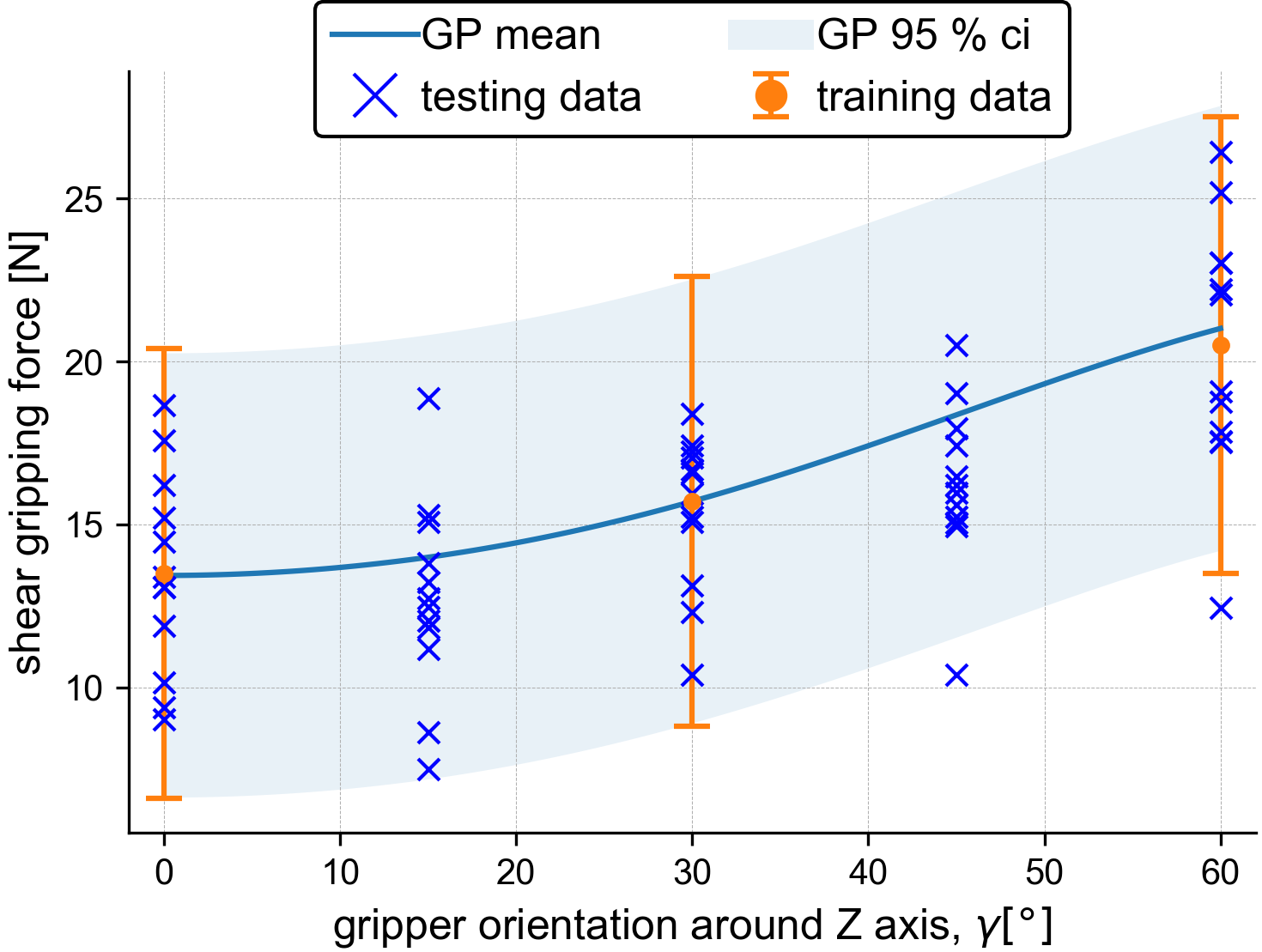}
    \caption{$\alpha = \ang{-15}, \beta = \ang{-15}, \lambda = 2.3$}
    \label{GP_z}
     \end{subfigure}
               \begin{subfigure}{0.24\textwidth}
         \centering
    \includegraphics[width=0.9\textwidth, clip]{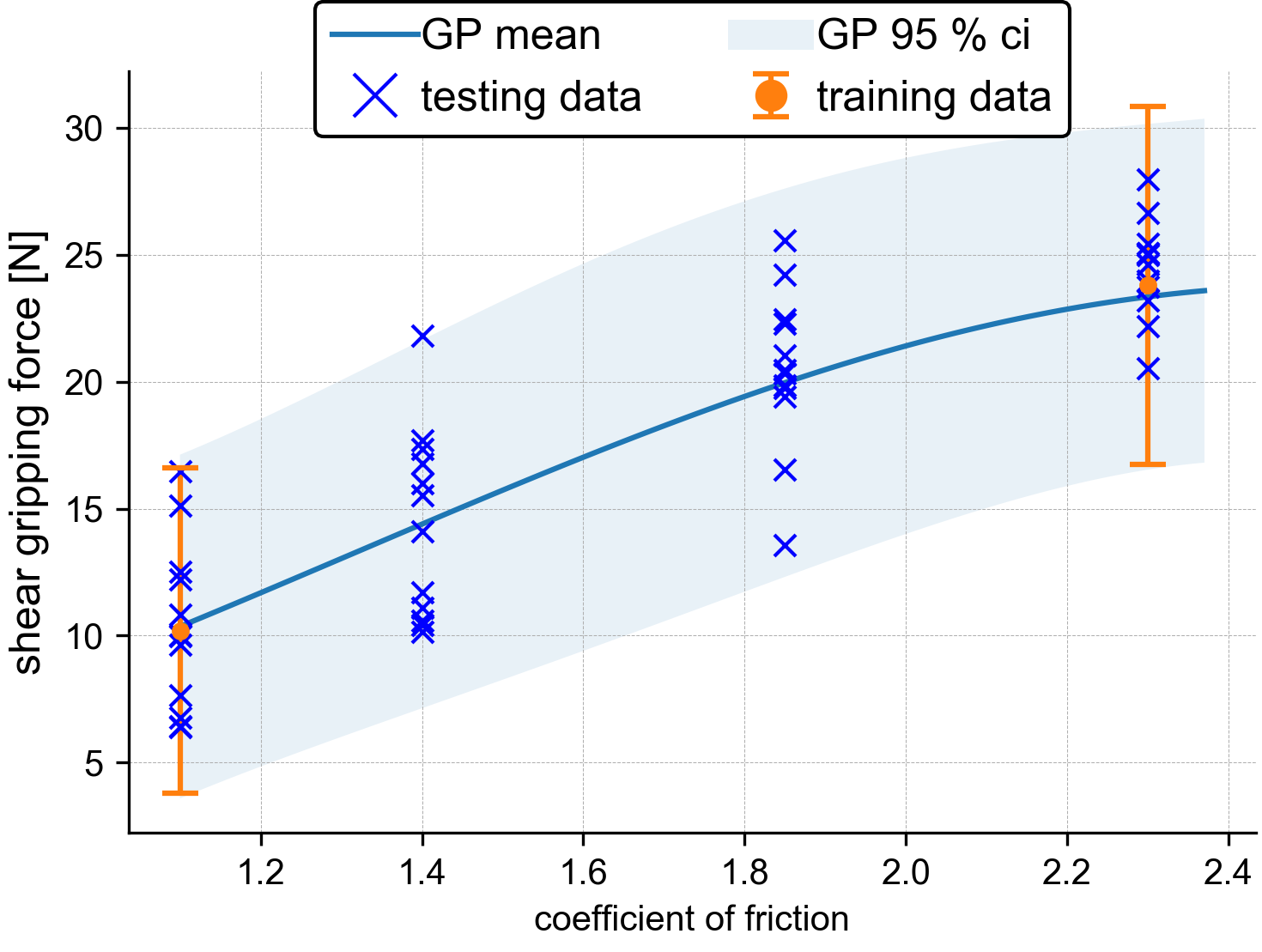}
    \caption{$\alpha = \ang{0}, \beta = \ang{0}, \gamma = 0$}
    \label{GP_friction}
     \end{subfigure}
     \caption{\textcolor{black}{The predicted maximum gripping force PDF from GP, the training data PDF, and the testing dataset}}
     \label{GP_result}
\end{figure} 

\textcolor{black}{To collect a dataset, maximum}
gripping forces \textcolor{black}{$\bm{f}^{g,m}$} were evaluated with a minimal normal force at varied orientations as \textcolor{black}{summarized in \tab{GP_dataset}.} \textcolor{black}{We collected 20 data sequences for every orientation as a training dataset and 10 data sequences as a testing dataset.}
% follows: \textcolor{black}{$\alpha,\beta=  \ang{-15}, \ \ang{0}, \   \ang{15}$ and $\gamma=\ang{0}, \ \ang{30}, \ \ang{60}$.} 
The coefficient of friction between spines and environments was measured by loading a constant mass on the gripper. A small activation force is necessary to compress spine springs and ensure \textcolor{black}{that the spines are} touching the wall, but \textcolor{black}{is} assumed to be negligible. \textcolor{black}{These} orientations were selected to cover possible gripper angles during regular wall climbing.  The gripper was fixed to a linear slider at \textcolor{black}{an} orientation and pulled by a force gauge 
% \textcolor{black}{for 20 times} 
on \textcolor{black}{36-grit and 80-grit} sandpapers \textcolor{black}{that are commonly used to emulate rough surface with microscopic asperities \cite{microspine},} as shown in \fig{experiment_gripper}. 
% \textcolor{black}{We verified that the mean of the gripping force did not change under various normal loads, which justify \eq{f_decomposition}.}
% The gripper force data was analyzed through GP with a squared exponential kernel with no bias. 
The GP hyperparameters \textcolor{black}{were} optimized using the \textcolor{black}{BFGS} algorithm \textcolor{black}{\cite{GPopt}.}
%
% \subsubsection{\textcolor{black}{Testing}}
\textcolor{black}{The obtained testing data with the predicted PDF of the maximum gripping force and the PDF of the training data is illustrated in \fig{GP_result}. The predicted maximum gripping force and the training data are displayed as a mean $\pm$ with a 95 \% confidence interval. Overall, we show that the GP prediction works well with different states.}

% \vspace{20pt}
\section{RISK-AWARE MOTION PLANNING}\label{risk_planning}
In this section, we present \textcolor{black}{a} complete risk-aware motion planning algorithm \textcolor{black}{formulated as \eq{obj}-\eq{constK}}. The objective of \textcolor{black}{our proposed} planner is to find the optimal trajectory for the Center of Mass (CoM) position, its orientation, the foot position, and the reaction force for each foot in order to arrive at the destination while satisfying constraints. Our proposed planner enables the robot to find feasible trajectories that consider risk from the grippers \textcolor{black}{under} various environments.

\textcolor{black}{We define one \emph{round} of movement made by a robot when its body and all of its limbs have moved onto the next footholds.}
Note that for each round, the planner investigates several critical instants between two postures with pre-defined gait as explained in detail in Section~\ref{results}.
% In \textcolor{black}{\eq{obj}-\eq{constK}}, 
At $j$-th round, $\Gamma$ is the decision variables that are given as:
\begin{equation}
    \begin{aligned}
\Gamma=\{\bm {p}_{i,j}, {\bm{P}}_{\operatorname{CoM},j}, {\bm \Theta}_{\operatorname{CoM},j}, {\bm \theta}_{i,j},\bm f^r_{i,j}, \textcolor{black}{\hat{\bm f}_{i,j}^{g,m}, \hat{\bm \Sigma}_{i,j}^{g,m}} \}
\end{aligned}
\end{equation}
where  $\bm {p}_{i,j}$ is the foot $i$ position, ${\bm{P}}_{\operatorname{CoM},j}$ is the position of the body, ${\bm \Theta}_{\operatorname{CoM},j}$ is the orientation of the body, ${\bm \theta}_{i,j}$ \textcolor{black}{are} the joint angles for the limb $i$, and
\begin{figure}
\begin{subequations}
\begin{flalign}
\underset{\Gamma}{\operatorname{minimize}}\  \color{black} \Psi_{tot}\label{obj}
\\
% \left[\Delta \bm P_{\min }, \Delta \bm \Theta_{\min }\right] \leq\left[\Delta \bm P^{\operatorname{CoM}}_{i}, \Delta \bm \Theta^{b}_i\right] \leq\left[\Delta \bm P_{\max }, \Delta  \bm \Theta_{\max }\right]
\text{s.t., for each round $j=1, \ldots, N$}\nonumber\\
\text{ and for each limb $i=1, \ldots, L$}\nonumber\\
 |\Delta \bm P_{\operatorname{CoM}}| \leq\Delta \bm P_{\operatorname{Th} }\label{constA}&& \color{black}\operatorname{(linear\ stride)}
\\
|\Delta \bm \Theta_{\operatorname{CoM}}| \leq\Delta  \bm \Theta_{\operatorname{Th} }\label{constB}&& \color{black}\operatorname{(angular \ stride)}
\\
    |\Delta \bm p_{i}| \leq\Delta \bm p_{\operatorname{Th} }\label{constC}&& \color{black}\operatorname{(foot \ stride)}\\
\mathbf{\bm p}_{i,j} \in \mathcal{R}(\bm P_{\operatorname{CoM},j},\bm \Theta_{\operatorname{CoM},j}, \bm \theta_{i,j})\label{constD}&& \color{black}\operatorname{(kinematics)}
\\
  \color{black}  \mathbf{\bm p}_{i,j} \in \mathcal{T}\label{constE}&&
 \color{black}    \operatorname{(contact \ region)}
    % \mathbf{\bm p}_{i,j} \in  \mathcal{T}\label{constE}&&
    %  \operatorname{(feasible \ region)}
\\
    % -\bm f^r_{i,j}=\bm f^g_{i,j}+\bm f^e_{i,j}\label{constH}&&
    %  \operatorname{(reaction \ force)}\\
   \color{black} \sum_{i=1}^{L}\bm f^r_{i,j}+\bm{F}_{t o t}=\bm 0\label{constF}&&
  \color{black}   \operatorname{(force\ eqm)}
\\
\color{black}\sum_{i=1}^{L}\left(\bm{p}_{i,j} \times \bm f^r_{i,j}\right)+\bm{M}_{t o t}=\bm 0\label{constG}&&
  \color{black}   \operatorname{(moment\ eqm)}
\\
    \bm \tau_{i,j}=\bm J\left(\bm \theta_{i,j}\right)^{\top}  \color{black} \bm f^r_{i,j}\label{constI}&&
    \color{black} \operatorname{(joint \ torque)}
\\
    \left\|\bm \tau_{i,j}\right\|_{2} \leq  \bm \tau_{\operatorname{Th}}\label{constJ}&&
     \color{black} \operatorname{(torque \ limit)}
\\
    \bm{f}^r_{i,j} \in \mathcal{F}\left(\textcolor{black}{\lambda}_{i,j}(\bm{p}_{i,j}), \bm{n}_{i,j}, \textcolor{black}{ \bm f^{g,m}_{i,j} }\right)\label{constK}&&
    \textcolor{black}{\operatorname{(friction \ cone)}}
    % \\
%     \bm f^l_{i,j}=-\bm K_{i}\left(\bm \delta_{\operatorname{ wall },i,j}- \bm \delta_{\operatorname{CoM},i,j}\right)\label{constDeform}
% \\
% \bm{ K_{i,j}=\left(\bm J\left(\bm \theta_{i,j}\right) k^{-1} \bm J\left(\bm \theta_{i,j}\right)^{\top}\right)^{-1}}\label{constStiff}
\end{flalign}
\end{subequations}
\end{figure}
\textcolor{black}{$\bm f^r_{i,j}$ is defined in Section~\ref{friction_cone}. In this study, $\bm f^{g,m}_{i,j}$ is treated as a random variable based on the model of GP, which follows $\bm f^{g,m}_{i,j} \sim N\left( \hat{\bm f}_{i,j}^{g,m}(s_{i,j}), \hat{\bm \Sigma}_{i,j}^{g,m}(s_{i,j})\right)$.}
Equation \eq{obj} is the cost function that depends on the \textcolor{black}{robot's state}. Equation \eq{constA}, \eq{constB}, and \eq{constC} bound the range of travel between rounds. Equation \eq{constD} represents the forward kinematics constraints. In \eq{constE}, it ensures that $\bm {p}_{i,j}$ is within the feasible terrain where the robot is able to put its limb. 
% Because $\bm f^g_{i,j}$ is the random variable, $\bm f^r_{i,j}$ becomes a random variable in \eq{constH}. 
% Therefore, $\bm f^r_{i,j}$ follows $\bm f^r_{i,j} \sim N\left( \hat{\bm f}_{i,j}^r(s_{i,j}), \hat{\bm \Sigma}_{i,j}^r(s_{i,j})\right)=N\left(- \hat{\bm f}_{i,j}^g(s_{i,j})-{\bm f}_{i,j}^e, \hat{\bm \Sigma}_{i,j}^g(s_{i,j})\right)$.
% 
In this paper, we assume that the robot generates a quasi-static motion. Hence, the planner has the static equilibrium constraints expressed by \eq{constF} and \eq{constG}, where $\bm{F}_{t o t}$ and $\bm{M}_{t o t}$ is the external force and moment, respectively. In this work, only gravity is considered as the external force. 
% Note that \eq{constF} and \eq{constG} are stochastic constraints since  $\bm f^r_{i,j}$ is the random variable. Hence, the mean of $\bm f^r_{i,j}$  is used in \eq{constF} and \eq{constG} to hold the constraints. 
Equation \eq{constI} and \eq{constJ} ensure \textcolor{black}{that} the motor torque is lower than the maximum motor torque where $J\left(\bm \theta_{i}\right)$ is a Jacobian matrix. The reaction force $\bm f^r_{i}$ is constrained by \eq{constK}, which describes the friction cone constraints to prevent the robot from slipping \textcolor{black}{where ${\lambda}_{i,j}(\bm{p}_{i,j})$ denotes the coefficient of friction at $\bm {p}_{i,j}$}. Note that this constraints \eq{constK} is also stochastic constraints due to  $\bm f^r_{i}$. Equation \eq{constK} can be converted into deterministic constraints, which is explained in Section~\ref{subsec:CC}.
% If a robot is position-controlled, the planner needs to add additional constraints to compute the control input to the motor to generate the planned forces, which are defined in \eq{spring_f}, \eq{bigKandJ}.

% Equation \eq{constDeform} and \eq{constStiff} are from \eq{spring_f}-\eq{deltaX}, which represent a force by a limb with its deflection. These two constraints are only for a position-controlled robot.

% When a robot is equipped with a gripper, a gripping force can be considered a force to resist slipping, resulting in that a robot is able to generate more reaction force $f^r_{i}(t)$ by keeping the same amount of $f^l_{i}(t)$ or decrease  $f^l_{i}(t)$ with the same amount of $f^r_{i}(t)$. Thus, introducing a gripper, a robot can climb on the more challenging environment or at least plan a more energy-efficient trajectory. This relation is expressed in \eq{constH}.

Compared to sampling-based approaches such as \textcolor{black}{RRT}, NLP \textcolor{black}{is able to formulate} relatively complicated constraints such as friction cone constraints \eq{constK}, which are typically difficult for the sampling-based approaches to handle \textcolor{black}{in terms of computation}. In addition, MICP approaches such as \cite{MIT_envelope}, \cite{robust_gait}, \cite{xuanMICP} \textcolor{black}{can increase the computation speed} by decoupling the pose state from wrench states. However, \textcolor{black}{they} potentially limit the \textcolor{black}{robot's mobility.} The robot may not choose the trajectory on the low friction terrain in case the planner first solves the pose problem and then solves the wrench problem later \textcolor{black}{since} the pose optimization problem does not consider the wrench information. \textcolor{black}{Although MICP can} plan the trajectories \textcolor{black}{considering both wrench and pose state simultaneously, it needs to sacrifice the accuracy by assuming an envelope approximation on bilinear terms \cite{MIT_envelope} or allow relatively expensive computation as the number of the integer variables increases, which is intractable for high degree-of-freedom (DoF) robots (e.g., our robot has 24 DoF). In contrast, NLP can simultaneously solve the trajectory reasoning both the pose and the wrench with relatively less computation \cite{ETHNLP}.}

\subsection{Deterministic Constraints}
Here, we explain two deterministic constraints \eq{constD}, \eq{constE},  that are not explicitly shown in \textcolor{black}{\eq{obj}-\eq{constK}}.
\subsubsection{\textcolor{black}{Kinematics}}
Forward kinematics \eq{constD} is \textcolor{black}{given} as:
\begin{equation}
    {\bm p}_{i,j} ={\bm R(\bm \Theta_{\operatorname{CoM},j})\bm p^b_{i,j}+\bm P_{\operatorname{CoM},j}}
\end{equation}
where $\bm R(\bm \Theta_{\operatorname{CoM}})$ is the rotation matrix from the world frame to the body frame, $\bm p^b_i$ is the foot position relative to the body frame. 
\subsubsection{Feasible Contact Regions}
We utilize NLP to formulate the planning algorithm so that any nonlinear terrain (i.e., non-flat terrain), such as tube and curve, can be directly described. 
% Obstacle avoidance can be realized by defining these constraints, which do not include the obstacle terrain.

If a robot traverses on the flat terrain, the footstep regions are convex polygons as follows:
\begin{equation}
    \bm{C}_{r} \bm{p}_{i,j} \leq \bm{D}_{r}
\end{equation}

\subsection{Chance Constraints}\label{subsec:CC}
Here, we show that the friction cone constraints in \eq{constK} can be expressed using chance constraints, which allow the planner to convert the stochastic constraints into deterministic constraints.

One key characteristic of robotic climbing is that climbing is a highly risky operation: a robot can easily fall without planning its motion correctly. \textcolor{black}{Hence}, it needs to restrict reaction forces using the friction cone constraints given as:
\begin{equation}
    \bm{n}_{i,j}^{\top}\bm{f}_{i,j}^{r}  \geq 0\label{df1}
\end{equation}
\begin{equation}
    \left\|\bm{f}_{i,j}^{r}-\left(\bm{n}_{i,j}^{\top}\bm{f}_{i,j}^{r}\right) \bm{n}_{i,j}^{\top}\right\|_{2} \leq\textcolor{black}{\lambda}_i\left(\bm{n}_{i,j}^{\top} \bm{f}_{i,j}^{r}\right)\textcolor{black}{+\bm{f}_{i,j}^{g,m}}\label{friction_pyramid}
\end{equation}
To decrease the computation of solving for the NLP solver, we simplify the \eq{friction_pyramid} by linearizing them as follows:
% To utilize the chance constraints explained later and to decrease the computation of solving for the NLP solver, we simplify the \eq{friction_pyramid} by linearizing them as follows:
\begin{equation}
    \left|\bm{\zeta}_{i,j}^{\top}\bm{f}_{i,j}^{r} \right| \leq {\lambda}_{i}\left(\bm{n}_{i}^{\top}\bm{f}_{i,j}^{r}  \right)\textcolor{black}{+\bm{f}_{i,j}^{g,m}}\label{df2}
\end{equation}
\begin{equation}
    \left|\bm{\xi}_{i,j}^{\top}\bm{f}_{i,j}^{r} \right| \leq {\lambda}_{i}\left(\bm{n}_{i}^{\top}\bm{f}_{i,j}^{r}  \right)\textcolor{black}{+\bm{f}_{i,j}^{g,m}}\label{df3}
\end{equation}
where $\bm{\zeta}_{i,j}$, $\bm{\xi}_{i,j}$ are any tangential direction vectors on the wall plane. 
% Note that \eq{df1}, \eq{df2}, \eq{df3} are not deterministic constraints anymore since a gripping force by a spine gripper is stochastic, resulting in that we need to treat the adhesion force as a random variable which follows a multivariate Gaussian  distribution defined as:
% \begin{equation}
%     \bm f^r \sim N\left(-\bm f^e-\hat{\bm f}^g, \hat{\bm \Sigma}^g\right)
% \end{equation}
% By dealing with $\bm f^g$ as a random variable, the whole optimization becomes stochastic optimization problem and constraints \eq{constF}, \eq{constG}, \eq{constH},  \eq{constK}  are chance constraints. Hence, in  \eq{constF}, \eq{constG}, the expectation is taken on the reaction force term.

% \textcolor{black}{Regarding \eq{constK} formulated as \eq{df1}, \eq{df2}, \eq{df3}, we can define the feasible region $\mathcal{F}$ as an intersection of $M$ linear inequality constraints as:}

Regarding \eq{constK} formulated as \eq{df1}, \eq{df2}, \eq{df3}, we rearrange the equations and the \textcolor{black}{joint} chance constraint is given by:\textcolor{black}{
\begin{equation}
    \textcolor{black}{\operatorname{Pr}\left(\bigwedge_{j=1, \ldots, N}\bigwedge_{k=1, \ldots, M}\bm \alpha_{i,j}^{k\top}\bm f_{i,j}^{g,m}\leq\beta_{i,j}^{k}\right) \geq 1 - \Delta_{\operatorname{}}}\label{jointrisk}
\end{equation}
where $\bm \alpha_{i,j}^{k}$ are \textcolor{black}{coefficient} vectors, and $\beta_{i,j}^{k}$ are \textcolor{black}{coefficient} scalars that consist of the convex polytopes defined in \eq{df1}, \eq{df2}, \eq{df3}. In \eq{jointrisk}, $M$ denotes the number of constraints defining the polytopes.}
\textcolor{black}{$\Delta_{\operatorname{}}$} is the user-defined violation probability, where the probability of violating constraints is  under the \textcolor{black}{$\Delta_{\operatorname{}}$}. \textcolor{black}{We can regard $\Delta$ as relating to the likelihood that gripper slip will be responsible for the failure of the robot.} For example, if \textcolor{black}{$\Delta_{\operatorname{}}$} is high, the planner can explore a larger space because the feasible region expands in optimization. As a result, the robot tends to plan a trajectory \textcolor{black}{with a high violation probability}  by assuming that the gripper generates enough force. For a robotic climbing task, these chance constraints enable the robot to 
% \textcolor{black}{perform} energy-efficient 
\textcolor{black}{perform} challenging \textcolor{black}{motions} that would be infeasible without considering the gripping force.   In contrast, if \textcolor{black}{$\Delta_{\operatorname{}}$} is small, the planner tends to generate more conservative motions because the robot assumes that the gripper does not output enough force to support the weight of the robot.

 \textcolor{black}{Imposing \eq{jointrisk} is computationally intractable. Thus, using Boole's inequality, Blackmore \cite{blackmore2}, showed that the feasible solution to \eq{jointrisk} is the feasible solution to the following equations:
\begin{equation}
    \textcolor{black}{\operatorname{Pr}\left(\bm \alpha_{i,j}^{k\top}\bm f_{i,j}^{g,m}\leq\beta_{i,j}^k\right) \geq 1 - \Delta_{j,k}}\label{individual_risk}
\end{equation}
\begin{equation}
    \sum_{j=1}^{N} \sum_{k=1}^{M} \Delta_{j,k} \leq \Delta_{\operatorname{}}\label{sdelta_non}
\end{equation}
for all $j=1, \ldots, N, \quad k=1, \ldots, M$.} The violation probability for each \textcolor{black}{constraint} per round  \textcolor{black}{$\Delta_{j,k}$} is constrained in \eq{sdelta_non}, in order not to exceed the given \textcolor{black}{$\Delta_{\operatorname{}}$}. \textcolor{black}{Because non-uniform risk allocation \eq{sdelta_non} is also computationally expensive \cite{risk_allocation},} we use  the following relation:
\begin{equation}
     \color{black}   {\Delta_{j,k}} = \frac{\Delta_{\operatorname{}}}{NM}\label{sdelta}
\end{equation}

% Here, $\bm{\alpha}_{i, j}^{\top} \bm{f}_{i, j}^{r}$ is a multivariate Gaussian distribution such that $\bm{\alpha}_{i, j}^{\top} \bm{f}_{i, j}^{r} \sim N\left(\bm{\alpha}_{i, j}^{\top}\left(-\bm f_{i,j}^e-\hat{\bm f}_{i,j}^g\right), \bm{\alpha}_{i, j}^{\top}\hat{\bm \Sigma}^g\bm{\alpha}_{i, j}\right)= N\left(\bm{\alpha}_{i, j}^{\top}\hat{\bm f}_{i,j}^r, \bm{\alpha}_{i, j}^{\top}\hat{\bm \Sigma}^r\bm{\alpha}_{i, j}\right)$. It can be then converted into a deterministic constraint as given by:

\textcolor{black}{$\bm{\alpha}_{i, j}^{k\top} \bm{f}_{i, j}^{g,m}$} is a multivariate Gaussian distribution such that \textcolor{black}{$\bm{\alpha}_{i, j}^{k\top} \bm{f}_{i, j}^{g,m} \sim N\left(\bm{\alpha}_{i, j}^{k\top}\hat{\bm f}_{i,j}^{g,m}, \bm{\alpha}_{i, j, k\top}\hat{\bm \Sigma}^{g,m}_{i,j}\bm{\alpha}_{i, j}^k\right)$}. Thus, the stochastic constraints \eq{individual_risk} can be then converted into a deterministic constraint as given by:
\begin{equation}
\begin{split}
   \color{black}    \operatorname{Pr}\left(\bm \alpha_{i,j}^{k\top}\bm f_{i,j}^{g,m}\leq\beta_{i,j}^k\right)&=  \color{black}\Phi\left(\frac{ \beta_{i,j}^k-\bm{\alpha}_{i, j}^{k\top}\hat{\bm f}_{i,j}^{g,m}}{\sqrt{\bm{\alpha}_{i, j}^{k\top}\hat{\bm \Sigma}^{g,m}_{i,j}\bm{\alpha}_{i, j}^{k}}}\right)
\\
     &  \color{black}\geq 1-\Delta_{j,k}
     \end{split}
\end{equation}
where $\Phi$ is the cumulative distribution function of the standard normal distribution. It can be transformed further as follows:
\begin{equation}
  \color{black} \bm{\alpha}_{i, j}^{k\top}\hat{\bm f}_{i,j}^{g,m}  +  \sqrt{\bm{\alpha}_{i, j}^{k\top}\hat{\bm \Sigma}^{g,m}_{i,j}\bm{\alpha}_{i, j}^k}\Phi^{-1}\left( 1-\Delta_{j,k} \right) \leq \beta_{i,j}^k \label{full_chanceconstrain}
\end{equation}
where $\Phi^{-1}$ is the inverse \textcolor{black}{function} of $\Phi$.

\subsection{Cost Function}
The cost function consists of intermediate costs and a terminal cost. In this work, the target mission is to arrive at the destination. Thus, the terminal cost is the distance from the position of the last \textcolor{black}{pose} to the destination. 
\begin{equation}
 \textcolor{black}{\Psi_{\operatorname{D}}}=\left({\bm{q}}_N-{\bm{q}}_{D}\right)^{\top} {\mathbf{W}}_{\operatorname{D}}\left({\bm{q}}_N-{\bm{q}}_{d}\right)
\end{equation}
where ${\mathbf{W}}_{\operatorname{D}}$ is the weighting matrix and ${\bm{q}}_N=\left[{\bm{p}}_{1,N}, \ldots, {\bm{p}}_{L,N}\right]$ while ${\bm{q}}_{d}$ is the configuration at the destination.
The intermediate costs restrict a large amount of shifting in terms of linear and rotational motion of a body and the foot position as follows:
\begin{equation}
    \begin{aligned}
&\textcolor{black}{\Psi_{\operatorname{BPos}}}=\Delta {\bm{P}}_{\operatorname{CoM}}^{\top} {\mathbf{W}}_{\operatorname{BPos}} \Delta {\bm{P}}_{\operatorname{CoM}}\\
&\textcolor{black}{\Psi_{\operatorname{Foot}}}=\sum_{i=1}^{L} \Delta {\bm{p}}_{i}^{\top} {\mathbf{W}}_{\operatorname{Foot}} \Delta {\bm{p}}_{i}\\
&\textcolor{black}{\Psi_{\operatorname{BRot}}}=\Delta {\Theta}_{\operatorname{CoM}}^{\top} \mathbf{{W}_{\operatorname{BRot}}}  \Delta {\Theta}_{\operatorname{CoM}}
\end{aligned}
\end{equation}
where ${\mathbf{W}}_{\operatorname{BPos}}$, ${\mathbf{W}}_{\operatorname{Foot}}$, and ${\mathbf{W}}_{\operatorname{BRot}}$ are the weighting matrix.
\subsection{Two Step Optimization for a Position-Controlled Robot}
Although \textcolor{black}{our proposed} motion planner works for any limbed robot, there is a drawback \textcolor{black}{for} a position-controlled robot \textcolor{black}{when} wall-climbing. For the position-controlled robot, it is necessary to compute how much ${\bm \delta}_{\operatorname{wall}}$ is necessary to generate the planned reaction forces. Therefore, the planner needs to include  additional constraints from \eq{spring_f}, \eq{bigKandJ} to realize the planned trajectory. However, we observed that the nonlinear solver has \textcolor{black}{a} numerical issue with \eq{bigKandJ}, so it is intractable for the solver to solve \textcolor{black}{our proposed} NLP in \textcolor{black}{\eq{obj}-\eq{constK}} with \eq{spring_f}, \eq{bigKandJ}. To avoid this problem, we decouple the optimization problem into two-step problems shown in \textcolor{black}{\eq{new_obj}-\eq{risk_allocation_eq} and \eq{nobj}-\eq{nconstStiff}}: 
\begin{subequations}
\begin{flalign}
% \underset{\Gamma}{\operatorname{minimize}}\ J_{\operatorname{D}}+\sum_{j=1}^{N-1}\left(J_{\operatorname{BodyPos}}+J_{\operatorname{Foot}}+J_{\operatorname{BodyRot}}\right)\label{newobj}
\underset{\Gamma}{\operatorname{minimize}} \ \color{black} \Psi_{\operatorname{D}}+\sum_{j=1}^{N-1}(\Psi_{\operatorname{BPos}}+\Psi_{\operatorname{Foot}}+\Psi_{\operatorname{BRot}})\label{new_obj}
\\
    \text{s.t.} |\bm P_{\operatorname{CoM},j+1}-\bm P_{\operatorname{CoM},j}| \leq\Delta \bm P_{\operatorname{Th}}\label{nconstA}
    % && \operatorname{(fr)}
\\
    | \bm \Theta_{\operatorname{CoM},j+1}-\bm \Theta_{\operatorname{CoM},j}| \leq\Delta  \bm \Theta_{\operatorname{Th} }\label{nconstB}
\\
    |\bm p_{i,j+1}-\bm p_{i,j}| \leq\Delta \bm p_{\operatorname{Th} }\label{nconstC}\\
{\bm p}_{i,j} ={\bm R(\bm \Theta_{\operatorname{CoM},j})\bm p^b_{i,j}+\bm P_{\operatorname{CoM},j}}\label{nconstD}
\\
    \bm{C}_{r} \bm{p}_{i,j} \leq \bm{D}_{r}\label{nconstE}
\\
\color{black} \sum_{i=1}^{L}{\bm{f}}_{i, j}^{r}+\bm{F}_{t o t}=\bm 0\label{nconstF}
\\
\color{black} \sum_{i=1}^{L}\left(\bm{p}_{i,j} \times  {\bm{f}}_{i, j}^{r}\right)+\bm{M}_{t o t}=\bm 0\label{nconstG}
\\
    \bm \tau_{i,j}=\bm J\left(\bm \theta_{i,j}\right)^{\top}  \textcolor{black}{\bm f^r_{i,j}}\label{nconstI}
\\
    \left\|\bm \tau_{i,j}\right\|_{2} \leq \Delta \bm \tau\label{nconstJ}
\\
 \color{black} \bm{\alpha}_{i, j}^{k\top}\hat{\bm f}_{i,j}^{g,m}  +  \sqrt{\bm{\alpha}_{i, j}^{k\top}\hat{\bm \Sigma}^{g,m}_{i,j}\bm{\alpha}_{i, j}^k}\Phi^{-1}\left( 1-\Delta_{j,k} \right) \leq \beta_{i,j}^k
%   \eq{full_chanceconstrain}
   \label{nconstK}
  \\
 \textcolor{black}{{\Delta_{j,k}} = \frac{\Delta_{\operatorname{}}}{NM}\label{risk_allocation_eq}}
\end{flalign}
\end{subequations}
\begin{subequations}
\begin{align}
{\operatorname{find}}\ \bm \delta_{\operatorname{ wall },i,j},  \bm \delta_{\operatorname{CoM},i,j}\label{nobj}
\\
\text{s.t.}  \left\|\bm \delta_{\operatorname{ wall },i,j}\right\|_{2} \leq  \bm \delta_{\operatorname{Th,  wall },i,j}\label{delta_abs}
\\
  \textcolor{black}{\bm f^r_{i,j}}=\bm K_{i}\left(\bm \delta_{\operatorname{ wall },i,j}- \bm \delta_{\operatorname{CoM},i,j}\right)\label{nconstDeform}
\\
\bm{ K_{i,j}=\left(\bm J\left(\bm \theta_{i,j}\right) k^{-1} \bm J\left(\bm \theta_{i,j}\right)^{\top}\right)^{-1}}\label{nconstStiff}
\end{align}
\end{subequations}
the first planner is in charge of the pose and the reaction force of the robot, and the second planner finds $\bm \delta_{\operatorname{ wall },i,j},  \bm \delta_{\operatorname{CoM},i,j}$, which are the control inputs to a position-controlled robot. In \textcolor{black}{\eq{nobj}-\eq{nconstStiff}}, the constraint \eq{delta_abs} ensures that $\bm \delta_{\operatorname{ wall },i,j}$ is bounded under a certain threshold.

We argue that this decoupling is reasonable because the first planner solves the \textit{"essential"} problem (e.g., How much reaction force is necessary? What is the footstep trajectory?) to plan the \textcolor{black}{force and pose} trajectory. The second planner \textcolor{black}{only computes} the control input to the motors, \textcolor{black}{and} it does not have a significant effect on the entire motion planning. As explained, if the robot is force controlled, the planner does not need to consider  \eq{spring_f}, \eq{bigKandJ}. As a result, the second optimization is not necessary for a force-controlled robot, and the whole motion is planned only based on the first optimization problem.    
%what is the better way to explain better.
% In conclusion, the motion planning algorithm  shown in \textcolor{black}{\eq{obj}-\eq{constK}} is decoupled into the two step optimization problem as shown in \fig{1st_optimization}, \fig{2nd_optimization}.

% \begin{figure}
%     \caption{ Nonlinear programming formulation for a pose and a force planning}
%     \label{1st_optimization}
% \end{figure}
% \begin{figure}
%     \caption{Convex programming formulation for a position-controlled robot}
%     \label{2nd_optimization}
% \end{figure}

\section{RESULTS}\label{results}

In this section, we \textcolor{black}{evaluate \textcolor{black}{our proposed} planner by testing the robot's performance in three different tasks}: energy-efficient climbing, climbing on non-uniform terrains, and climbing with \textcolor{black}{a tripod} gait. 

We utilize Ipopt solver \cite{IPOPT} to solve the \textcolor{black}{planning} problem on an Intel Core i7-8750H machine. \textcolor{black}{The derivative of constraints are provided by CasAdi \cite{casadi}. The optimizer is initialized with the default configuration of the robot (\fig{low_friction_avoid_a_3d}, bottom configuration), and the specifications of the computation for Section~\ref{exp2_section} is summarized in \tab{computation}.} 

\begin{table}[t]
    \caption{\textcolor{black}{NLP specifications for climbing on non-uniform walls}}
    \centering
    \begin{tabular}{c|c|c|c}
      \# of rounds $N$  & Variables & Constraints & Average T-solve (Ipopt)\\
         \hline
         1 & 1744 & 779 & 0.4 minutes\\
         2 & 3937 & 1680 & 6 minutes\\
         4 & 11761 & 4994 & 16 minutes\\
         7 & 23479 & 9965 & 248 minutes
    \end{tabular}
    \label{computation}
\end{table}

We implement the \textcolor{black}{results of our} proposed planning algorithm \textcolor{black}{(i.e., the motion plan)}, on a six-limbed robot, each limb of which has three DoF. \textcolor{black}{Each joint uses pairs of} Dynamixel MX-106 motors, providing a maximum torque at 27 Nm. The robot is equipped with \textcolor{black}{a} battery, computer, and IMU. The robot runs a PID loop to regulate its body orientation. No other sensor is used to control its linear position. The robot weighs 11.5 kg. The width of the robot's body is 442 mm while its height at its standing state is  180 mm.   In each experiment, the robot climbs between two walls at a distance of 1200 mm, where the wall is covered with sandpapers of different grit size to adjust the coefficient of friction. \textcolor{black}{All hardware demonstrations can be viewed in the accompanied video\footnote{\textcolor{black}{ Video of hardware experiments: https://youtu.be/ZDqvf1J4nS4}}.}

\subsection{Energy Efficient Planning}
% 1. deterministic VS proposed planning with different possibility in one single wall
% 2. proposed planning with different possibility on still one single map, but the wall has different sandpapers 
% 3. obstacle avoidance: maybe a robot will take the more challenging motion because it is necessary to avoid obstacles (infeasible vs feasible)

The objective of this task is to \textcolor{black}{assess} the consumed energy of climbing with two different violation probabilities. 
\textcolor{black}{While the robot can grip the wall with a low violation probability (e.g., $\Delta_{\operatorname{}}=0.0005$), there is a disadvantage of consuming more energy.}
% During climbing, the robot can grip very hard to the wall, resulting in a lower failure chance ($\Delta_{\operatorname{}}=0.001$), with more consumed energy.
On the other hand, the robot may \textcolor{black}{perform} an energy-efficient motion with a higher violation probability (e.g., \textcolor{black}{$\Delta_{\operatorname{}}=0.1$}). Here, we set $N=7$, $\textcolor{black}{M=6}$ to compute \textcolor{black}{$\Delta_{j,k}$. To} show the trade-off between the consumed energy and the violation probability, we let the robot climb on the walls with one leg gait where the robot first lifts its right front limb, puts it on the next position, pushes its body up, lifts its right middle limb, and so on. 
Within each round, the planner investigates 12 critical instants for one leg gait: 6 instants after the robot lifts one limb, and 6 instants after the robot \textcolor{black}{places the limb on the next position and} pushes its body. The planner solves the optimization problem for these 12 instants. 
We measure the current $I_{i,t}$ and the voltage $V_{i,t}$ of each limb $i$ online when the robot climbs on the wall covered by the \textcolor{black}{36-grit} sandpapers with one leg gait and estimated the power per one limb at every sampling time $t$. The power $P_{i,t}$ is estimated as follows:
\begin{equation}
    P_{i,t}=V_{i,t} \times I_{i,t}
\end{equation}
% In the experiment, the robot lifts the right side limb from the front, middle, and back, and then it lifts the left side leg from the back, middle, and front. 

\begin{figure}[t]
    \begin{subfigure}{0.5\textwidth}
    \centering
\includegraphics[width=0.7\textwidth, clip]{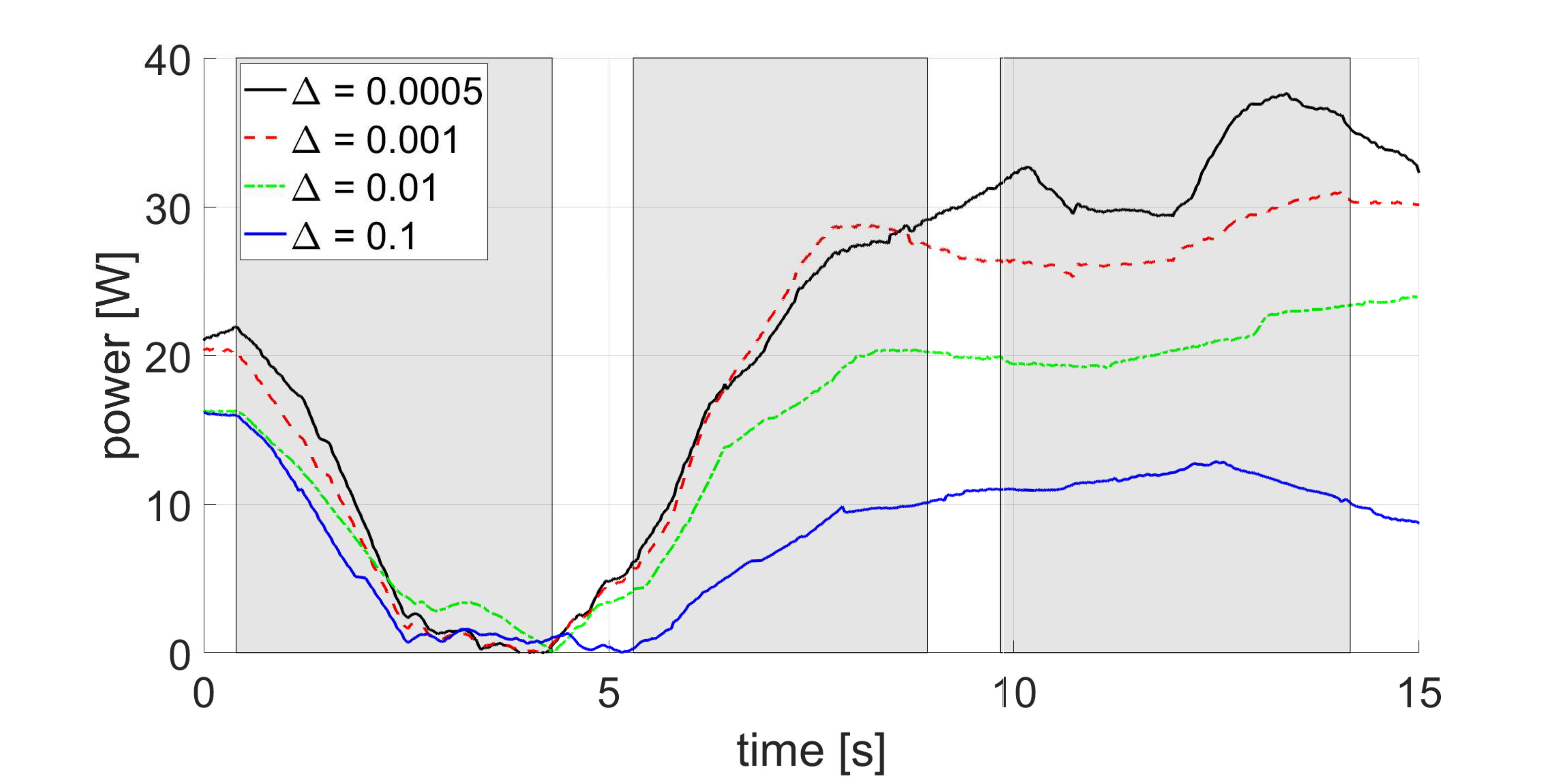}
    \caption{Time history of the consumed power for right front limb}
    \label{leg0}
    \end{subfigure}
    \\
     \begin{subfigure}{0.5\textwidth}
         \centering
    \includegraphics[width=0.7\textwidth, clip]{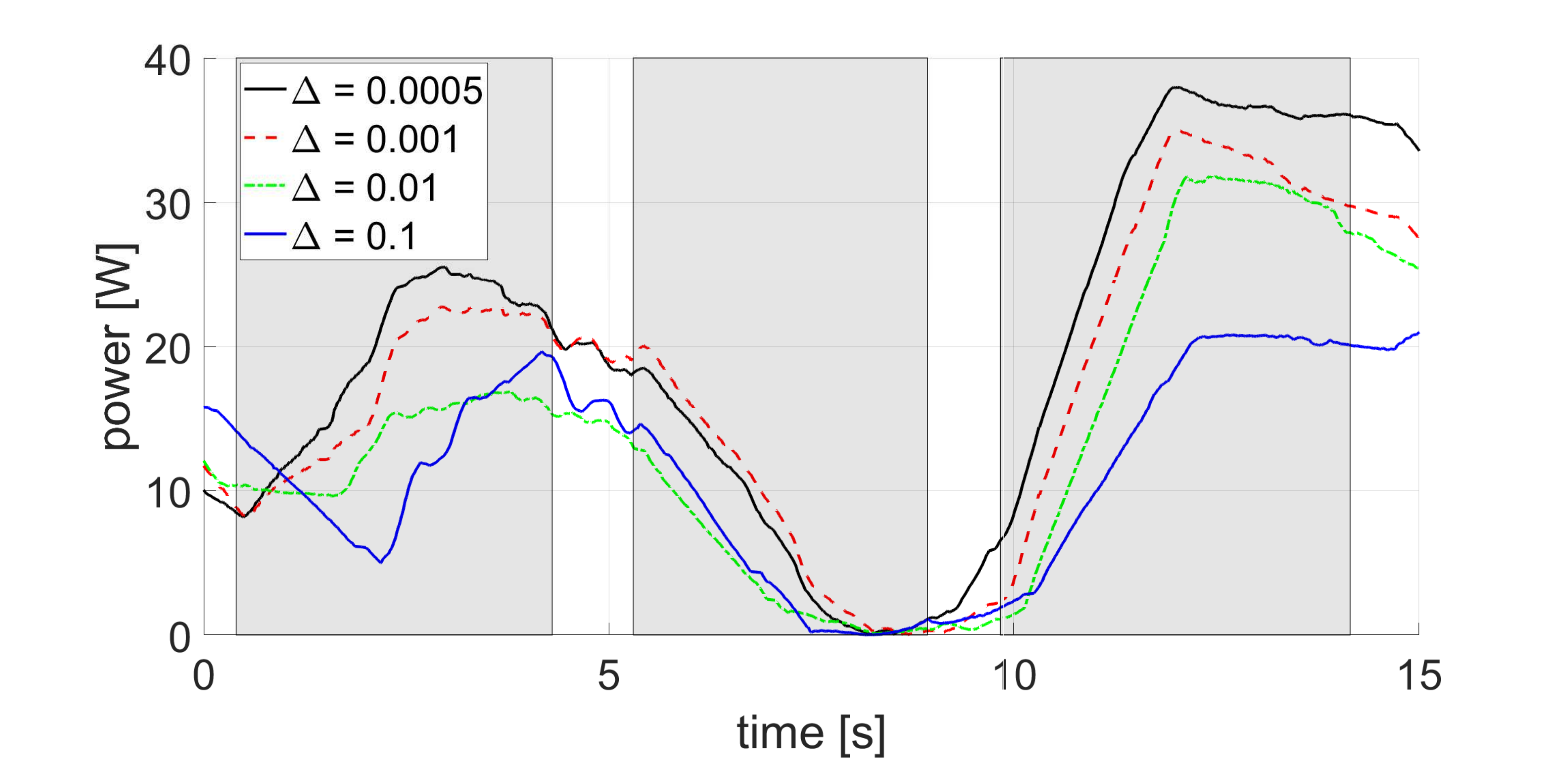}
    \caption{Time history of the consumed power for right middle limb}
    \label{leg1}
     \end{subfigure}
    %       \begin{subfigure}{0.5\textwidth}
    %      \centering
    % \includegraphics[width=0.
    % \caption{Time history of the consumed power for right back limb}
    % \label{leg2}
    %  \end{subfigure}
     \caption{Time history of the consumed power under the different violation probabilities. The shaded regions are when the robot lifts a specific limb and puts it on the next position, and white regions are when the robot pushes its body up. The figure shows that the consumed power of a particular limb decreases when the limb is in the air, while it increases when the limb is on the wall to generate the normal force on the wall.}\label{case1}
\end{figure}

\begin{figure}
    \centering
    \includegraphics[width=0.235\textwidth, clip]{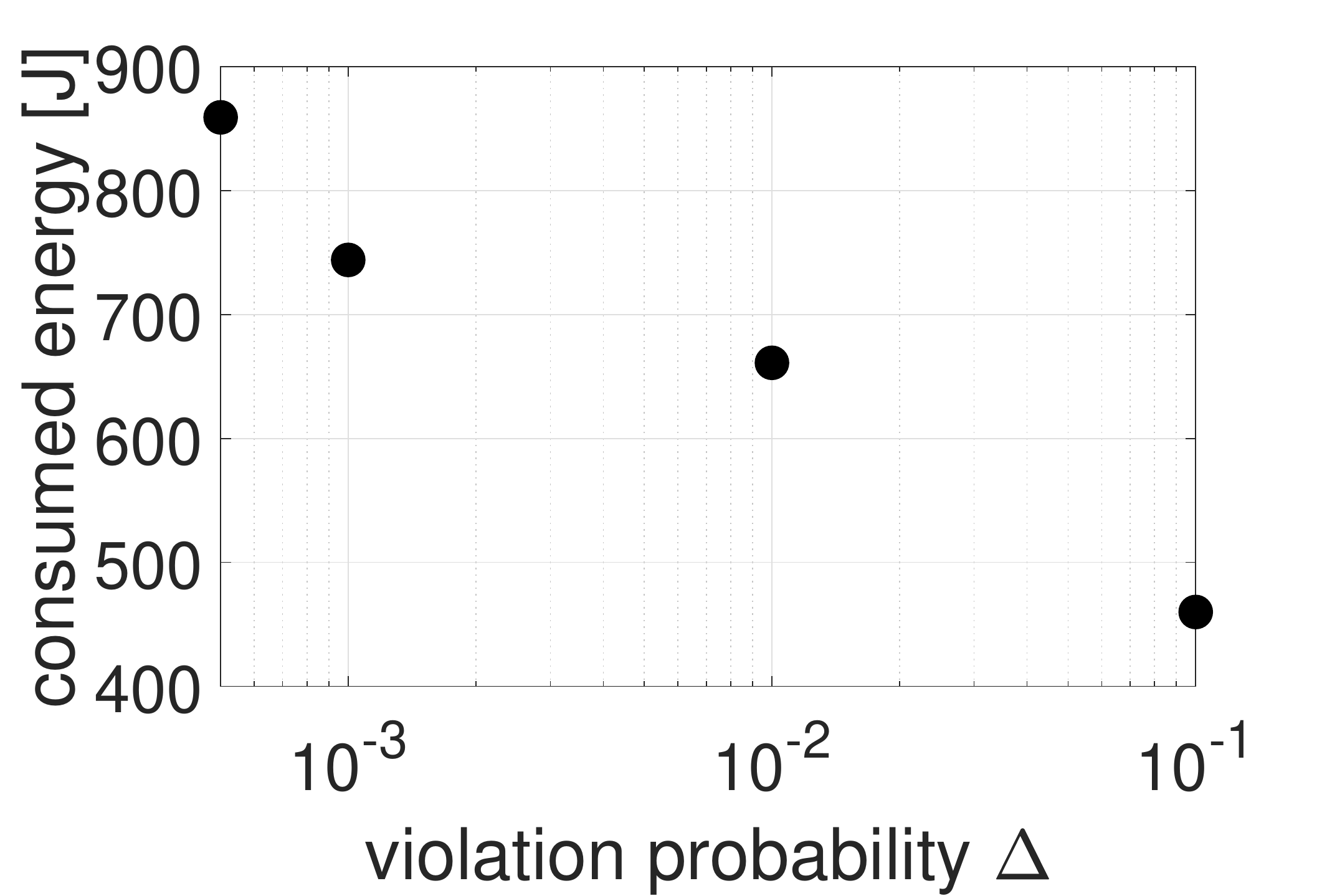}
    \caption{\textcolor{black}{ Consumed energy with different $\Delta_{\operatorname{}}$ during $t=0-15$ s}}
    \label{energy}
\end{figure}

\textcolor{black}{We plot the consumed power for two consecutive limbs from the hardware experiment in \fig{case1}. \fig{case1} shows that the consumed power of a limb decreases when the limb is in the air while the other limbs increase the consumed power to increase the reaction force.}
  Furthermore, the robot consumes \textcolor{black}{more} power with smaller \textcolor{black}{$\Delta_{\operatorname{}}$}, which means that the robot needs to push the wall to increase $f^r$. In contrast, if  \textcolor{black}{$\Delta_{\operatorname{}}=0.1$}, the solution requires less power, but has a larger probability of slipping. 
%   \textcolor{red}{We can also regard this comparison as the relation between the robust planner based on a deterministic gripping force with a worst-case uncertainty and the risk-bounded planner based on a PDF of a gripping force.}
%   This demonstrates the trade-off between energy \textcolor{red}{consumption} and the violation probability. 
\textcolor{black}{In \fig{energy}, the total consumed energy from these limbs was calculated by integrating their power over time spent climbing.}
%   \fig{energy} shows that the total consumed energy from these three limbs by integrating the power of these three limbs with the spent time of the climbing. 
  In our robot, the robot could decrease the energy by \textcolor{black}{46.5} \% under \textcolor{black}{ $\Delta_{\operatorname{}}=0.1$} compared with the energy under  \textcolor{black}{$\Delta_{\operatorname{}}=0.0005$.}


\subsection{Climbing on Non-uniform Walls}\label{exp2_section}
\begin{figure}[!t]
         \centering
    \includegraphics[width=0.5\textwidth, clip]{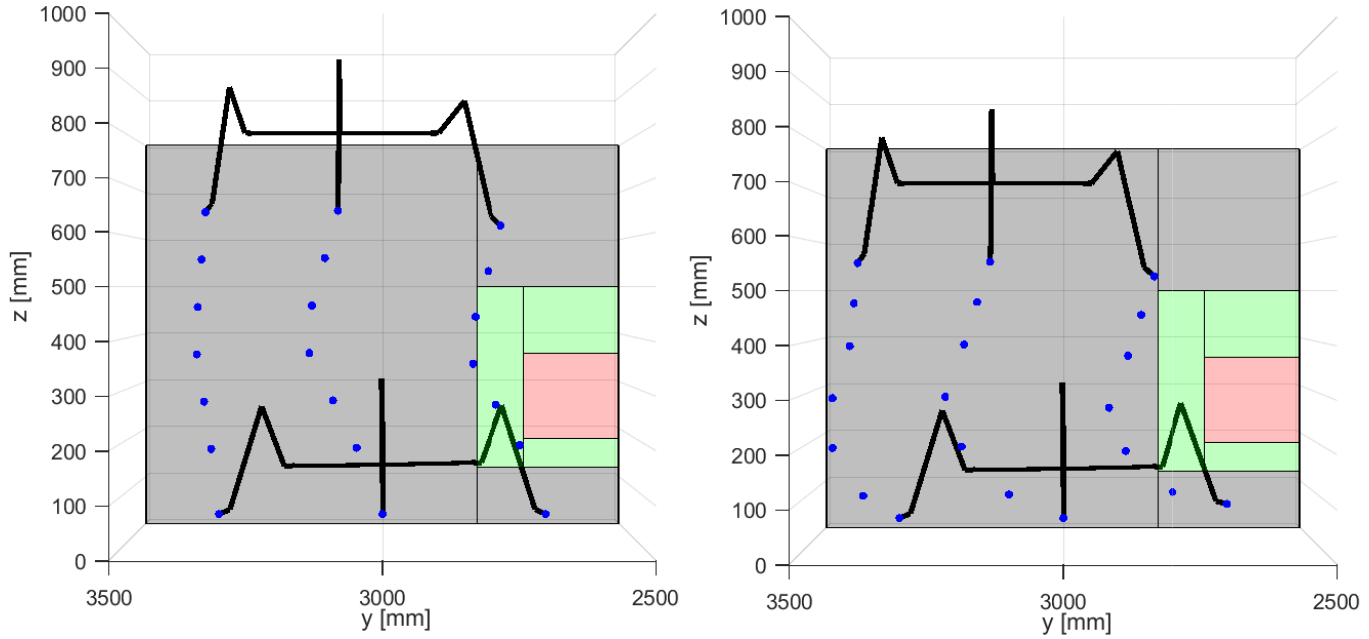}
         \caption{Side view of planned footsteps on non-uniform walls under: left $\Delta_{\operatorname{}} = 0.1$, right $\Delta_{\operatorname{}} = 0.001$. In \textcolor{black}{the} left \textcolor{black}{panel}, the robot puts its feet on low and high friction terrain by taking a high risk bound. In \textcolor{black}{the} right \textcolor{black}{panel}, the robot puts its feet \textcolor{black}{only} on high friction terrain.}\label{case222}
\end{figure}

% aggressive put green area still avoid red
% conservative avoids both green and red area
% Risk -> less length path to arrive at the destination
% conservative -> avoid low friction env
% experiment -> the robot avoids the low friction area but at 2nd round it puts its leg
This scenario demonstrates that the robot designs different trajectories under the different violation probability to climb on walls with varying coefficients of friction. The planned trajectories are shown in \fig{case222}. In this example, the robot climbs between two walls where the terrain shown in black is covered by \textcolor{black}{36-grit} sandpapers ($\lambda=2.3$), the terrain shown in green is covered by \textcolor{black}{80-grit} sandpapers ($\lambda=1.1$), and the terrain shown in red is covered by the material with $\lambda=0$ as shown in \fig{case222}. The varying coefficients of friction are modeled by a parabola function, which encourages the solver to converge on a solution. In \textcolor{black}{the} left \textcolor{black}{panel of} \fig{case222}, the violation probability \textcolor{black}{$\Delta_{\operatorname{}}$} is 0.1 while in \textcolor{black}{the} right \textcolor{black}{panel}, the violation probability  \textcolor{black}{$\Delta_{\operatorname{}}$} is 0.001 for $\textcolor{black}{M=6}$ and $N=7$. 

\begin{figure}[t]
    % \begin{subfigure}{0.5\textwidth}
    \centering
\includegraphics[width=0.44\textwidth, clip]{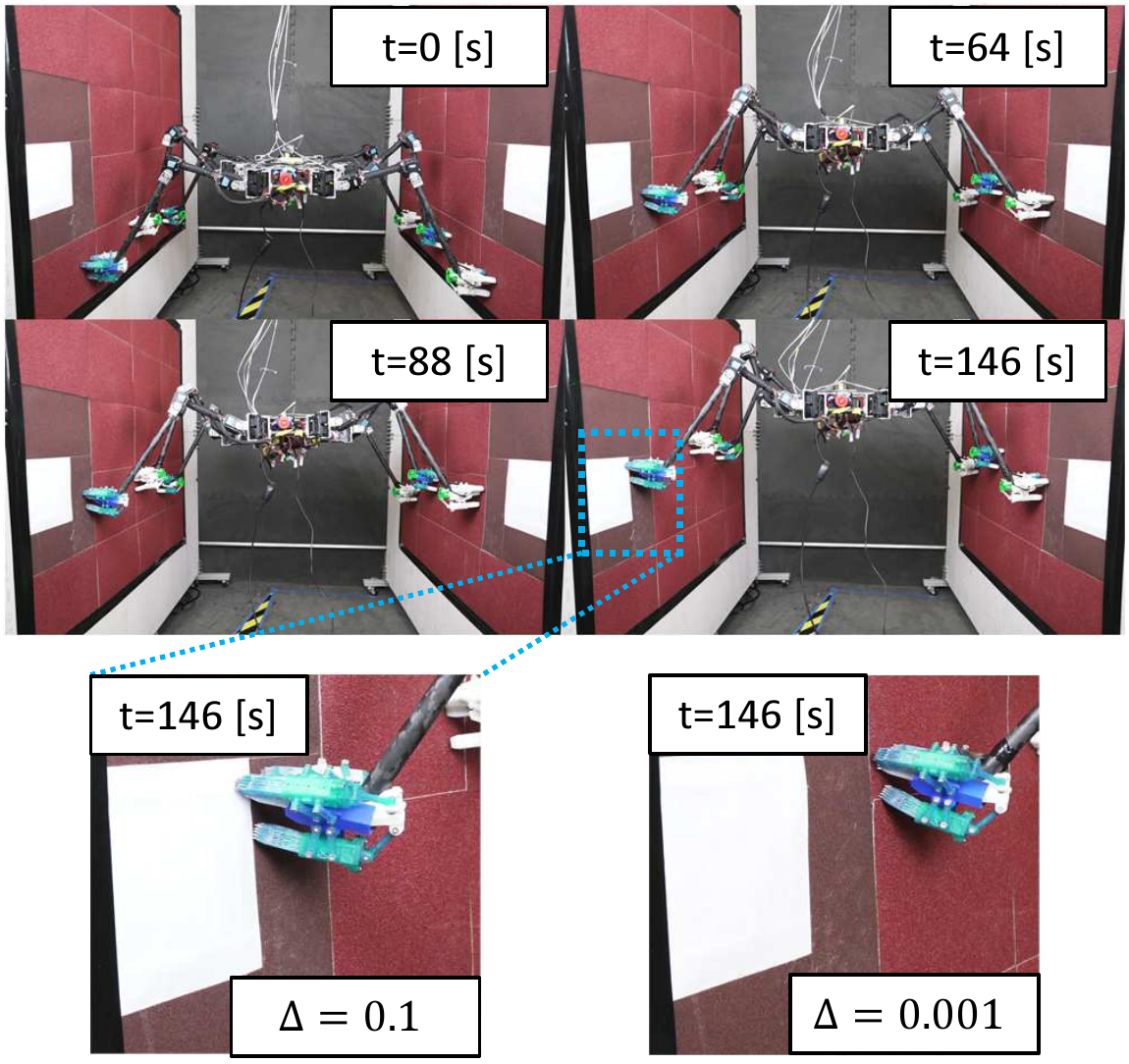}
    % \caption{$\Delta_{\operatorname{}}=0.1$ }
    % \label{case2_agg_exp}
    % \end{subfigure}\\
    %  \begin{subfigure}{0.5\textwidth}
        %  \centering
    % \includegraphics[width=0.75\textwidth, clip
    % \caption{$\Delta_{\operatorname{}}=0.001$ }
    % \label{case2_con_exp}
    %  \end{subfigure}
     \caption{Snapshots of climbing experiments on non-uniform walls under the different violation probabilities}\label{case2}
\end{figure}

\textcolor{black}{The left panel of} \fig{case222} illustrates that the robot avoids the red area (zero friction) and puts its foot mostly in the black area (high friction), but sometimes also in the green area (low friction) to minimize the trajectory length. In the right \textcolor{black}{panel of} \fig{case222}, the violation probability is decreased, and the robot footsteps completely remain inside the high friction area. As a result, \textcolor{black}{our proposed} NLP-based planner operates the pose and forces together and makes a trade-off between a shorter but more risky trajectory and a longer but safer trajectory. This cannot be achieved if the planner decouples the footstep and force planning, such as in \cite{xuanMICP}. 
\textcolor{black}{\fig{case2} shows the trajectory with higher risk bound $\Delta_{\operatorname{}}=0.1$ and compares the foot location at $t=146$ with the foot location with $\Delta_{\operatorname{}}=0.001$ in the hardware experiment.}
% Hardware experiment is shown in \fig{case2} where \fig{case2_agg_exp} shows the trajectory with higher risk bound $\Delta_{\operatorname{}}=0.1$, while \fig{case2_con_exp} shows the trajectory with lower risk bound $\Delta_{\operatorname{}}=0.01$. 
We notice that at $t=146$ s, the foot touches the white area where the coefficient of friction is 0, which never happened \textcolor{black}{with $\Delta_{\operatorname{}}=0.001$}. Since the robot only controls its body orientation based on IMU feedback and does not control its linear position, the implemented trajectory does not strictly follow the planned one. We observe that lower risk bound is beneficial in this situation to avoid failure since it compensates for the tracking error by the imperfect controller.
% by the imperfect controller.

\subsection{Climbing with \textcolor{black}{Less Stable}  Gait: Tripod Gait}

   \begin{figure}[t]
    \begin{subfigure}{0.24\textwidth}
    \centering
\includegraphics[width=1\textwidth, clip]{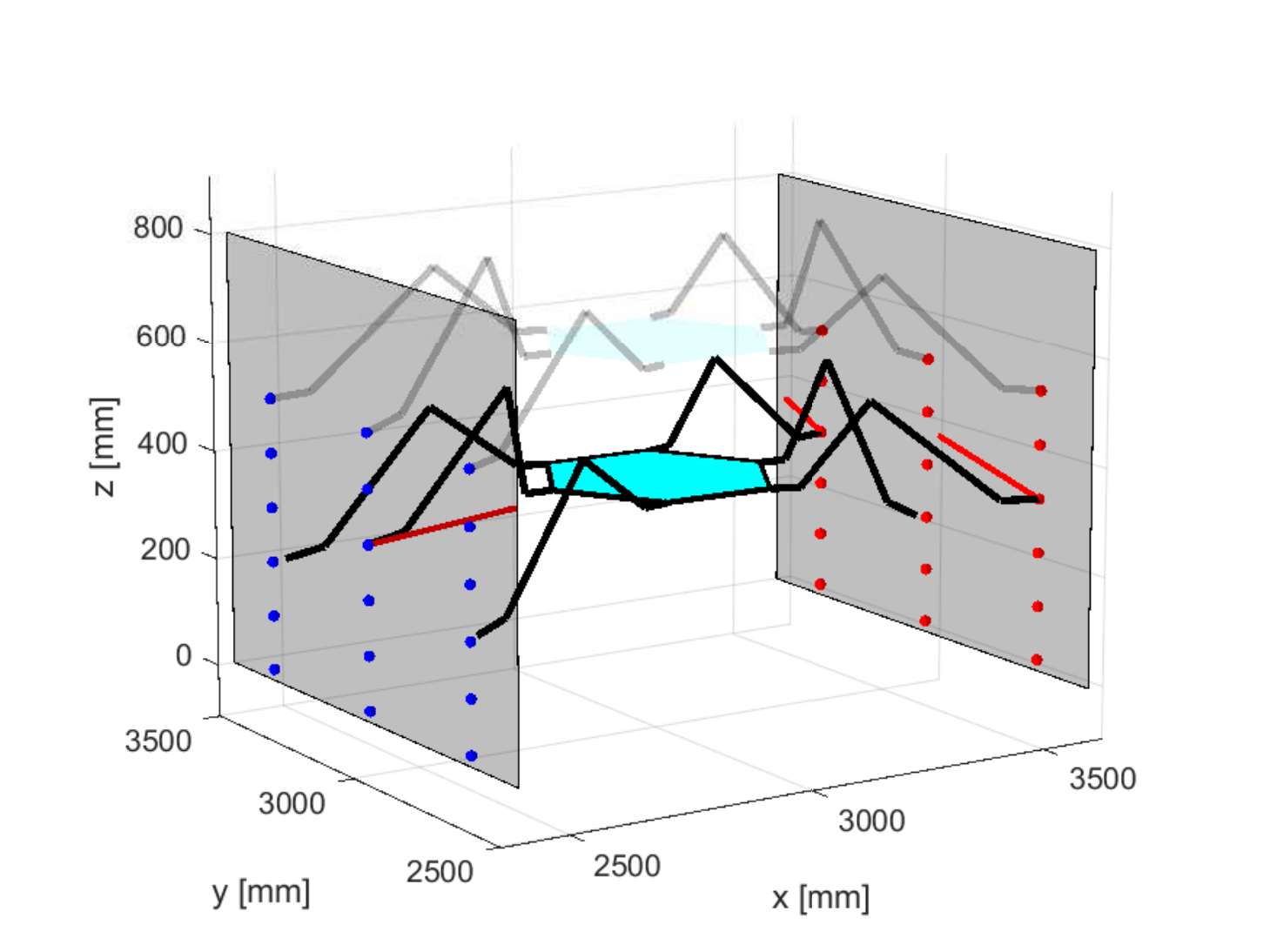}
    % \caption{}
    % \label{case3}
    \end{subfigure}
     \begin{subfigure}{0.24\textwidth}
         \centering
    \includegraphics[width=\textwidth, clip]{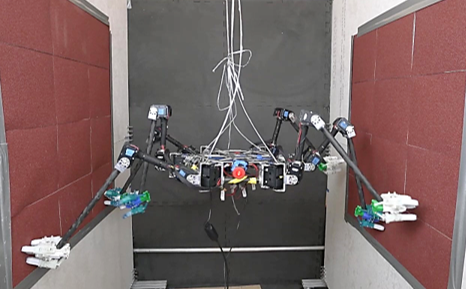}
    % \caption{}
    % \label{case3_exp2}
     \end{subfigure}
     \caption{Climbing with tripod gait. \textcolor{black}{Left: A planned trajectory of the tripod gait under $\Delta_{\operatorname{}}=0.4$. Red arrows indicate the reaction forces from the walls. Right: A snapshot of climbing experiments with the tripod gait under $\Delta_{\operatorname{}}=0.4$.}}\label{tripod}
\end{figure} 

% A planned trajectory of a tripod gait under $\Delta_{\operatorname{}}=0.4$. Red arrows indicate\\ the reaction forces from the walls. 
% Snapshots of climbing experiments with tripod gait under $\Delta_{\operatorname{}}=0.4$

In this scenario, we demonstrate that the robot can \textcolor{black}{conduct} a tripod gait, when it lifts three legs simultaneously, by setting \textcolor{black}{the} violation probability much higher. Before installing the gripper on the current six-limbed robot, it \textcolor{black}{was} almost infeasible to climb on the walls with the tripod gait because of the torque limits of the motors. With the grippers installed, \textcolor{black}{however,} the robot has a \textcolor{black}{greater} chance to climb on the walls with a  tripod gait. 
\textcolor{black}{If we set $\Delta=0$, the problem is infeasible since the constraints under the worst-case uncertainty are conservative. This result would be equivalent to the results of other robust algorithm such as \cite{autocar}, where the optimization-based robust approach with the worst-case uncertainty is proposed.}
% If we apply the \textcolor{black}{the optimization-based robust approach considering the worst-case uncertainty \cite{autocar}} to plan the trajectory, the planning problem is infeasible \textcolor{black}{since the constraints under worst-case uncertainty are conservative}. 
However, by utilizing the chance constraints and increasing the  violation probability, the planner generates a feasible solution. In our trial, we set  the violation probability \textcolor{black}{$\Delta_{\operatorname{}} =0.4$} for $\textcolor{black}{M=6}$ and $N=3$, and \textcolor{black}{allowed} the robot \textcolor{black}{to} climb on \textcolor{black}{a} wall covered by \textcolor{black}{36-grit} sandpapers. The planner investigates 4 critical instants: 2 instants after the robot lifts three limbs, and 2 instants after the robot \textcolor{black}{places them down and}  pushes its body. 
The planned trajectory is illustrated in \textcolor{black}{the left panel of} \fig{tripod}. As shown in \textcolor{black}{the right} under the condition, the robot succeeded in climbing on the walls with the tripod gait and \textcolor{black}{its} climbing velocity \textcolor{black}{was} 2.5 cm/s, which is three times faster than the one leg gait.
% Hence, the robot showed that it has the capability of climbing with tripod gait equipped with grippers by taking a large relative risk. However, we observed that the robot failed to climb more than 40 \% using tripod gait since the robot sometimes has a large amount of rotational sag-down, which is not modeled in this paper. Therefore, modeling that considers this rotational sag-down is for future research.
% During the experiment, we observed that the robot has sometimes large amount of rotational sag-down, which is not modeled in this paper. As for the challenging gait that has less stable region in the controller, this rotational sag-down causes 
% faster heavier. 

\section{CONCLUSION and FUTURE WORKS}\label{conclusion}
In this paper, we presented a motion planning algorithm for limbed robots with \textcolor{black}{stochastic gripping forces}. Our proposed planner exploits NLP to simultaneously plan a pose and force with guaranteed bounded risk. \textcolor{black}{Maximum} gripping forces are modeled as a Gaussian distribution by employing the GP, which provides the planner with the mean and the covariance information to formulate the chance constraints. We showed that under our planning framework, the robot demonstrates rich - sometimes drastically different - behaviors, including planning a risky but energy-efficient motion versus a safe but exhausting motion, avoiding danger zones like low friction environments, and choosing fast but \textcolor{black}{less stable} motions (i.e., a tripod gait) based on the different violation probabilities \textcolor{black}{$\Delta_{\operatorname{}}$} in hardware experiments. 

\textcolor{black}{The current limitation in this work is that the actual probability of failure is not strictly equal to pre-defined $\Delta_{\operatorname{}}$ because other sources of uncertainty exist, such as sensor noises. In future work, we will extend our planner to take into consideration these sources.}

\end{document}